\documentclass[10pt,twocolumn,letterpaper]{article}

\usepackage[pagenumbers]{wacv} 

\usepackage{algorithm}
\usepackage{algpseudocode}
\usepackage{amsmath}
\usepackage{amssymb}
\usepackage{multirow}

\definecolor{wacvblue}{rgb}{0.21,0.49,0.74}
\usepackage[pagebackref,breaklinks,colorlinks,allcolors=wacvblue]{hyperref}

\usepackage{tikz}
\usepackage{xcolor}

\title{
EgoNeMo: Transferable Map of Pedestrian Dynamics via\\ Egocentric LiDAR Scan
}

\author{
Azusa Sawada\\
NEC Corporation\\
{\tt\small swd02@nec.com}
\and
Allan Wang\\
Miraikan - The National Museum of Emerging Science and Innovation\\
{\tt\small allan.wang@jst.go.jp}
\and
Hideo Saito\\
Keio University\\
{\tt\small hs@keio.jp}
\and
Aaron Steinfeld\\
Carnegie Mellon University \\
{\tt\small steinfeld@cmu.edu}
 }

\begin{document}
\maketitle
\begin{tikzpicture}[remember picture, overlay]
  \node[anchor=north, font=\small\color{gray}\itshape, yshift=-1.5cm, xshift=-6cm] at (current page.north) {Preprint. Under review.};
\end{tikzpicture}

\begin{abstract}
This paper proposes a transferable Map of Dynamics (MoD) framework that generalizes to unknown environments using only egocentric 3D LiDAR point clouds to overcome the long-standing limitation of traditional MoD methods. While MoDs are essential for encoding human motion characteristics to enable accurate pedestrian trajectory prediction or safe robot navigation, traditional approaches suffer from site-specificity, requiring exhaustive trajectory accumulation at every new location. Extending recent advances in neural implicit modeling, our framework trains a continuous, LiDAR-based MoD estimator across diverse environments. To mitigate the inherent sparsity and temporal bias of real-world trajectory data, we introduce a position-balanced sampling strategy and a multi-task learning architecture that jointly predicts motion distributions and a spatial frequency score map. The latter is further augmented by visibility-aware losses to compensate for incomplete observation data. Comprehensive experiments demonstrate that our method effectively reconstructs underlying motion maps even in unknown locations from a single instantaneous LiDAR scan, despite highly sparse training data. Finally, we show that our improvements enhance the reliability of downstream trajectory prediction. 
\end{abstract}
    
\section{Introduction}\label{sec:intro}
In environments where humans and robots coexist and cooperate, site-specific typical motion patterns and human flow characteristics are crucial information for pedestrian trajectory prediction and safe, efficient path planning. As a framework for representing such dynamic spatial features, the Map of Dynamics (MoD) has been widely investigated~\cite{tomasz_survey23}. Since pedestrian motion at a specific location often exhibits multimodal characteristics, comprising a mix of heading directions and speeds, it is generally modeled as a probability distribution. Recent studies have increasingly adopted the Semi-Wrapped Gaussian Mixture Model (SWGMM) to explicitly represent the joint distribution of speed and angle. These expressive MoDs have demonstrated excellent performance in various downstream tasks, including long-term pedestrian trajectory prediction and robot navigation in crowded environments~\cite{zhu2023clifflhmp}.

\begin{figure}[t]
    \centering
    \includegraphics[width=0.99\linewidth, trim=0 40 70 0, clip]{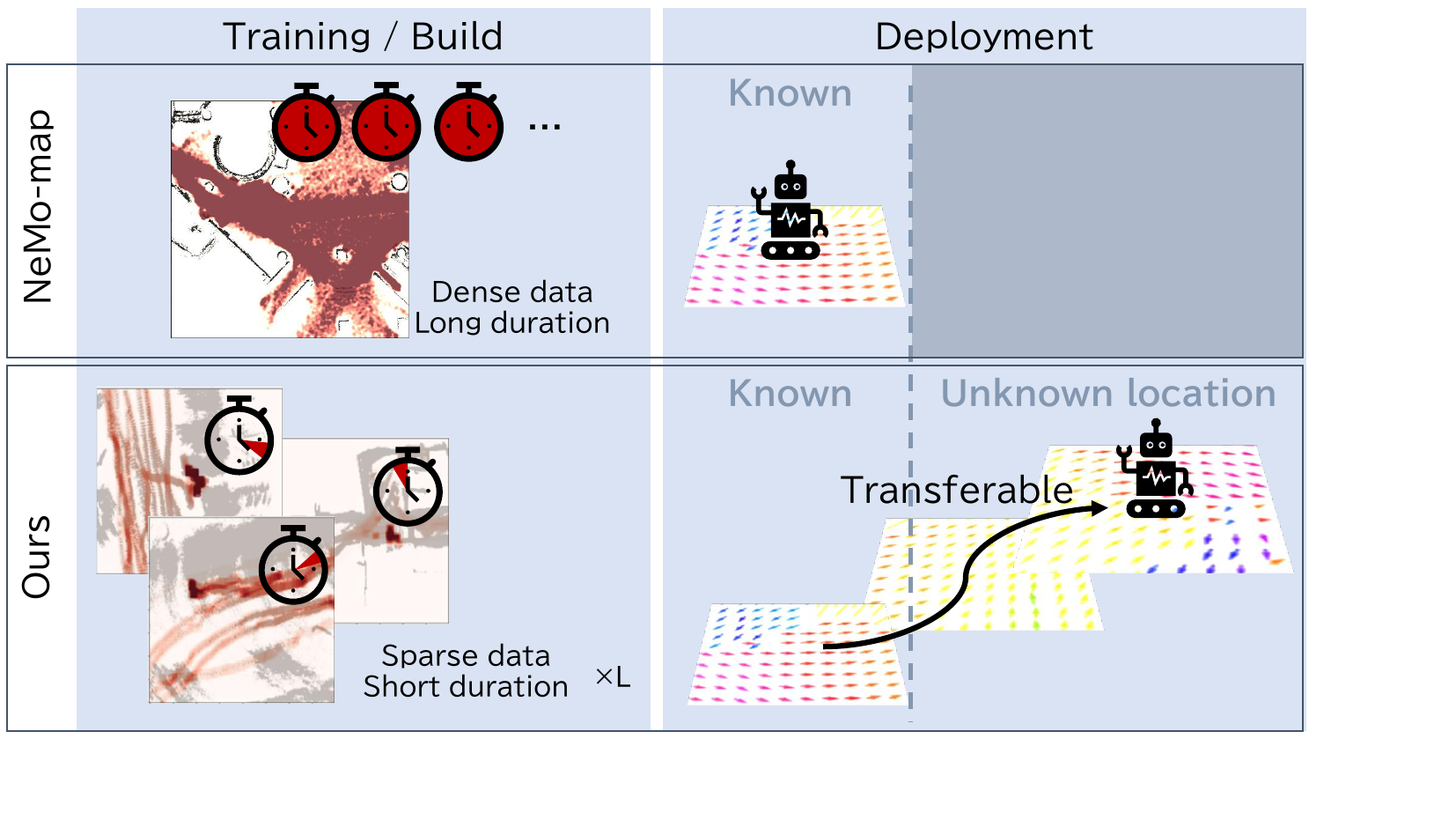}
    \caption{Concept of our framework compared to NeMo-map~\cite{zhu2026nemomap}. We train model across $L$ multiple locations to build transferable map of dynamics.}
    \label{fig:concept}
    \vspace{-5mm}
\end{figure}

However, deploying MoDs in real environments faces three major challenges. First, collecting a sufficient volume of pedestrian motion data across all possible positions is extremely time-consuming, and it becomes even more difficult in low-traffic environments. Second, conventional static mapping approaches, which rely on accumulated historical observation data, cannot immediately adapt to dynamic environmental changes, such as the temporary obstacles or layout alterations. Although sequential update methods~\cite{zhu2025cliffonline} can adapt to changes, it still requires time to observe and accumulate enough trajectories, which remains a critical bottleneck, preventing real-time environmental adaptation. Third, to generalize to unseen environments, some methods attempt to infer motion characteristics from static occupancy grids or floor plans; however, their representational capabilities are limited, for example, to transition probabilities between grid cells. Consequently, it remains difficult to use informative MoDs immediately at an unknown location when a robot first enters it.

To overcome these challenges, this paper proposes a MoD estimation framework that generalizes to unknown environments as illustrated in \cref{fig:concept}. 
The central idea is to extract the underlying relationships between motion dynamics and 3D geometry, including walls, entrances, and obstacles, by learning from trajectory datasets across multiple locations. Our approach takes LiDAR points from a mobile robot’s perspective as input 3D geometry, requiring only a single, instantaneous scan during deployment.
By extending recent neural implicit modeling by NeMo-map~\cite{zhu2026nemomap} to use LiDAR features, our framework estimates the MoD over continuous coordinates in the surrounding area. This eliminates the need to accumulate historical trajectory data at every location, enabling immediate prediction of continuous motion distributions in continuous space solely from the environmental observation.

The contributions of this paper are summarized as follows:
\begin{itemize}
\item We present the first transferable continuous MoD model, which instantly predicts motion dynamics in unknown environments directly from an egocentric LiDAR scan.
\item We introduce training methods to tackle the inherent sparsity in real-world data:
    (i) Joint training with frequency score prediction to enhance the generalization of motion outputs.
    (ii) Grid sampling to reduce the spatial bias arising from temporal continuity in the trajectory data.
    (iii) Visibility-guided frequency losses to handle incomplete and unlabeled frequency annotations.
\item We provide a deterministic downstream algorithm to consistently and efficiently predict trajectories from the estimated MoD without stochasticity.
\end{itemize}

Comprehensive experiments across both known and unknown locations in JRDB~\cite{martin2021jrdb,saadatnejad2023jrdb} demonstrate improved transferability over a baseline MoD extending NeMo-map and BFF~\cite{francesco2024bff}. 
We also show that the estimated MoDs provide reliable motion priors for trajectory predictions as a downstream task. 
These results demonstrate the feasibility of estimating MoDs directly from 3D geometry, opening a promising direction for transferable motion modeling in unseen environments.

\section{Related work}\label{sec:related}
Maps of Dynamics (MoDs) serve as environmental representations that store typical motion patterns characteristic of specific locations, effectively augmenting static geometric maps with statistical behavioral information~\cite{tomasz_survey23}. MoDs of human motion patterns are useful for robots to reason how to move compliantly in human-populated environments. MoDs typically fall into three categories: trajectory-based, which cluster complete paths~\cite{bennewitz2005learning}; occupancy-based, which treat dynamics as shifts in occupancy~\cite{wang2015modeling, wang2016building}; and velocity-based, which model local flow patterns directly from sparse velocity observations~\cite{kucner2017enabling}. The last variation, the Circular-Linear Flow Field map (CLiFF-map), has been influential because it can be built from a collection of velocity observations without complete or spatially dense trajectories. CLiFF-map represents multimodal flow using Semi-Wrapped Gaussian Mixture Models (SWGMMs) and can be directly used to predict future trajectories~\cite {zhu2023clifflhmp}. Some methods also incorporate temporal differences of motion patterns~\cite{zhi2019spatiotemporal, molina22exploration, zhu2026nemomap}.

These traditional MoDs rely on a spatial grid with a manually defined resolution, resulting in information loss and inconvenience. To overcome this limitation, ETMoD~\cite{shi2025etmod} automatically generates explainable and optimized cells and uses a Neural SDE to model temporal evolution.
Rheos~\cite{catalano2026rheos} utilizes meaningful locations defined by the 3D scene graph’s spatial hierarchy instead of a rigid grid. NeMo-map~\cite{zhu2026nemomap} directly maps spatio-temporal coordinates to SWGMM parameters by neural implicit representations, enabling continuous and resolution-independent queries across space and time. 

A fundamental limitation of traditional MoDs is their site-specificity; they describe only the environment in which they were trained and require extensive observation periods. To address this bottleneck, some researchers have developed transferable MoDs that generalize to unseen locations by leveraging the latent correlation between environmental geometry and human motion.
Early efforts explored using synthetic data to train networks in new settings or to predict macro-level activity indicators based on surrounding occupancy or floor plans~\cite {doellinger2018predicting, doellinger2019environment}. A significant milestone in this direction is the Bayesian Floor Field (BFF)~\cite{francesco2024bff}, which trains an occupancy-map-based CNN on real-world data to infer directional patterns in unseen environments even before any trajectory observations are available. By formulating the mapping process as a Bayesian inference problem, BFF can update a geometry-informed prior predicted by this CNN with on-site observations in high data efficiency.

While BFF demonstrated the transferability across environments, it remains inherently constrained by a grid-based, discrete transition model. In contrast, our work builds a transferable MoD model in a continuous, neural-implicit framework, similar to NeMo-map, to incorporate continuous motion modeling via a grid-free representation. Additionally, our framework operates on instantaneous, egocentric 3D LiDAR scans, which enable improved transferability and immediate response to environmental changes. 

\begin{figure*}[t]
    \centering
    \includegraphics[width=0.75\linewidth, trim=0 60 10 0, clip]{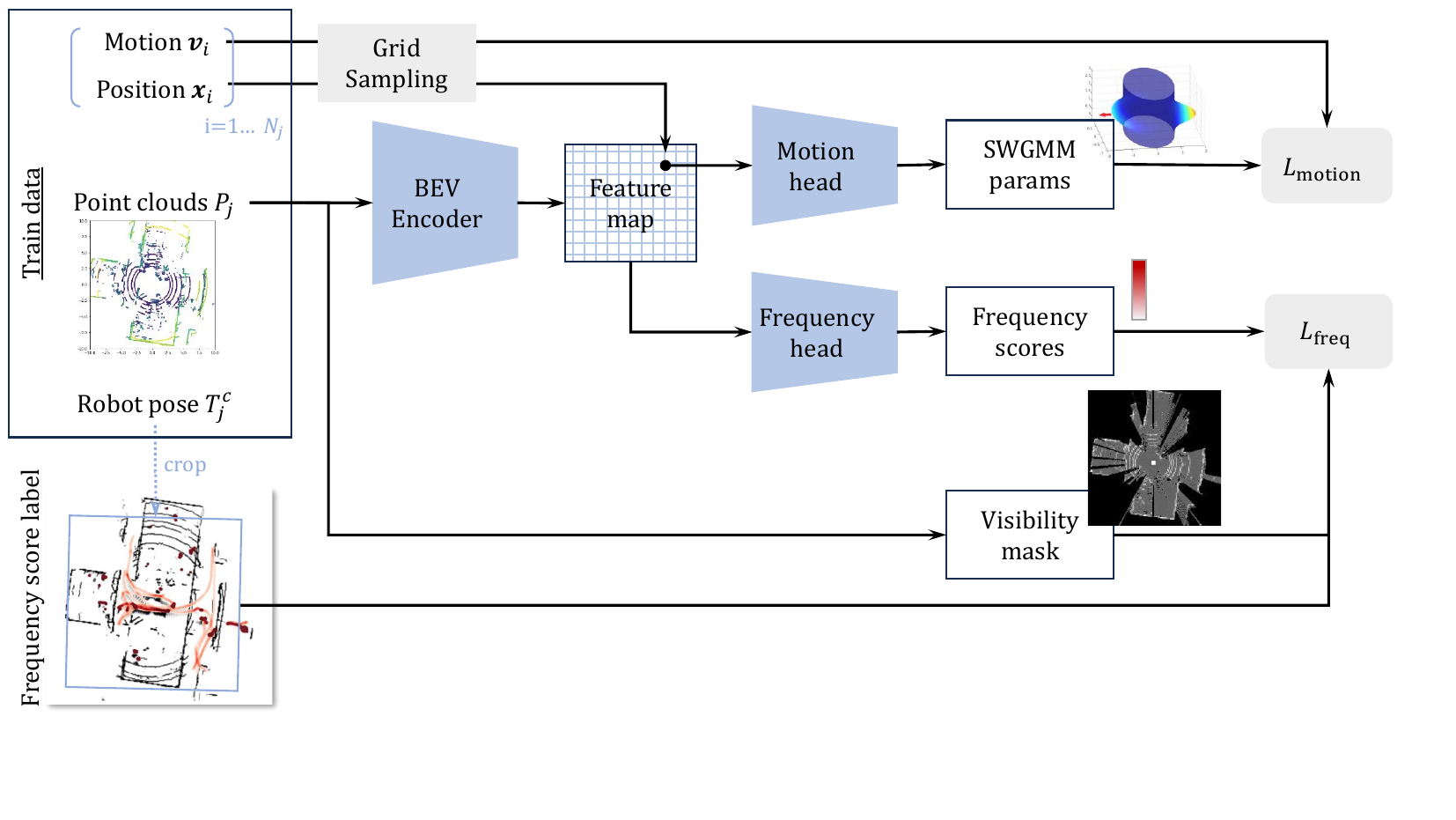}
    \caption{Overview of transferable MoD framework EgoNeMo. A neural network with a BEV encoder and two task heads is trained to predict Semi-wrapped Gaussian Mixture Model (SWGMM) parameters and a frequency score from LiDAR input. SWGMM image is from paper~\cite{zhu2026nemomap}}
    \label{fig:overview}
\end{figure*}

While existing transferable MoD models rely on static floor plans or binary occupancy maps, real-world robots typically perceive their surroundings as rich, high-fidelity 3D structural data. Methods for encoding such egocentric observations into 2D Bird's-Eye View (BEV) feature maps have been extensively studied in the autonomous driving literature for tasks such as BEV semantic segmentation and 3D object detection. These approaches are developed for sensor modalities: LiDAR-only~\cite{pointpillars2019}, RGB-only~\cite{philion2020lss,li2022bevformer,lu2025glss}, and LiDAR-RGB fusion models~\cite{liu2022bevfusion}. In this work, we adopt a lightweight LiDAR-only architecture to ensure both computational efficiency and direct geometric perception.

On the other hand, in popular trajectory prediction benchmarks, BEV RGB images are often available as they are collected for annotation, which has encouraged the development of methods that extract environmental priors from BEV images~\cite{mangalam2021waypoints, XIA2022csc, huang2024fullydecoupling, hu2025tscnet}. However, since such a top-down view is not always accessible in general environments, our framework is designed to operate solely on egocentric observations without the reliance on BEV images.

Overall, prior MoD approaches address different parts of the problem, but none provide a continuous and transferable MoD at the same time. BFF enables transfer to unseen environments via occupancy maps, but remains in a discrete grid transition model, whereas our method supports continuous spatial queries and continuous motion distributions. While NeMo-map provides continuous queries and distributions, it learns site-specific neural fields from on-site motion observations, whereas our model learns a transferable mapping from LiDAR geometry to MoD across locations.
\section{Method}\label{sec:method}
To enable immediate utilization of MoD in unknown environments, we propose a framework \textit{EgoNeMo} that predicts MoDs from egocentric environmental observations. During training, the model learns the relationship between LiDAR geometry and pedestrian motion from paired LiDAR scans and accumulated motion samples across multiple environments. At deployment, however, it requires only a single egocentric LiDAR scan and no trajectory observations in the target environment.


\Cref{fig:overview} illustrates the overview of our framework.
The model encodes 3D LiDAR points $P_j$ into a feature map $z_j=g(P_j)$ in a Bird’s-Eye View (BEV) coordinate as a geometric representation. Then, it maps the feature queried by arbitrary position $\textbf{x}_i$ to the Semi-Wrapped Gaussian Mixture Model (SWGMM)~\cite{Roy16SWGMM} via the motion head, following NeMo-map design~\cite{zhu2026nemomap}. The training data for this mapping consists of pairs of human motion samples  $\{(\textbf{x}_i, \textbf{v}_i)\}_{i=1}^{N_j}$  and LiDAR point clouds $\{P_j\}_{j=1}^M$ with a sensor (robot) pose $T_j^c$, where $\textbf{x}_i \in \mathbb{R}^2$ is 2D position and $\textbf{v}_i = (\rho_i, \theta_i)$ is velocity represented as speed $\rho_i \in \mathbb{R}^+$ and angle $\theta_i \in [0, 2\pi)$. We utilize train data from multiple locations to learn the common correlation between motion and geometry.

\begin{figure}[t]
    \centering
    \includegraphics[width=\linewidth]{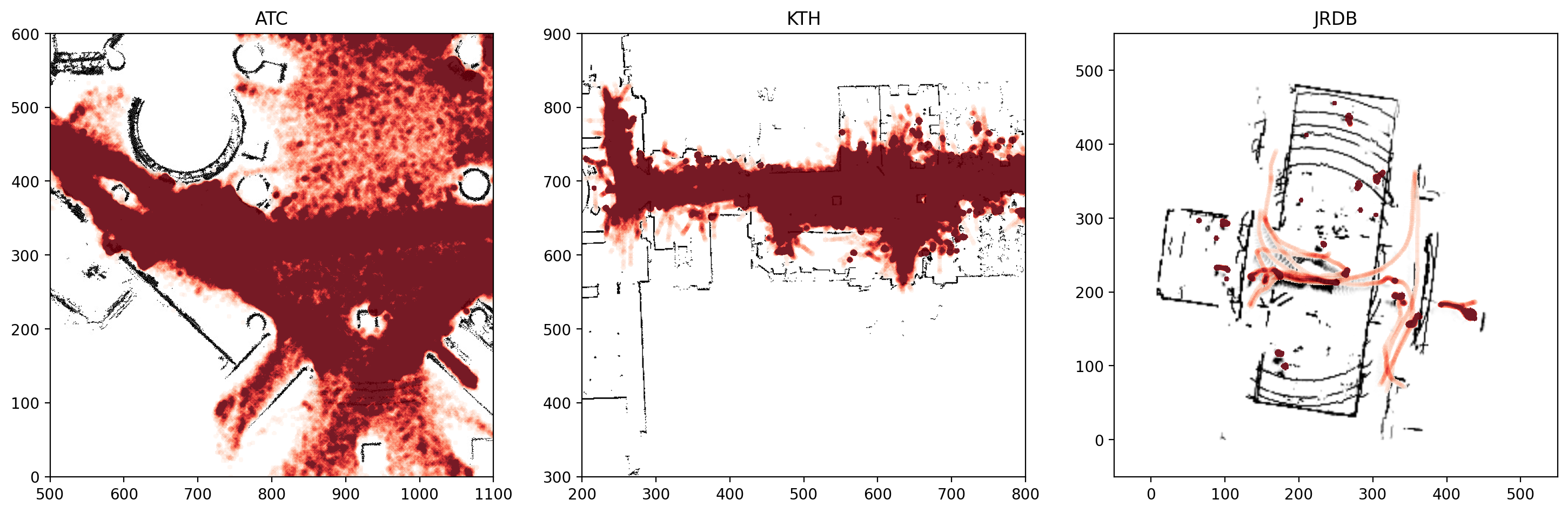}
    \caption{Spatial sparsity of ATC, KTH and JRDB. Red color shows position counts from all trajectories in a sequence.}
    \label{fig:sparsity}
\end{figure}

The core challenge is the severe sparsity of training data per location. Compared to the datasets frequently adopted in MoD literature~\cite{brscic2013person, Dondrup2015kth}, available multi-location datasets with egocentric LiDAR data have a significantly shorter duration (and lower density) of motion observations per location, resulting in severe spatial sparsity as illustrated in~\cref{fig:sparsity}. 
The improvements introduced to mitigate this challenge are explained in the following sections, along with the details of our model architecture, training process, and trajectory prediction algorithm.

\subsection{Model architecture}
We employ a PointPillars-based encoder~\cite{pointpillars2019} $g$ to extract a BEV feature map from LiDAR points after cropping into a fixed-size surrounding area ($20$ m $\times 20$ m throughout in this paper). 
For a given query coordinate $\textbf{x}_j$, it performs bilinear interpolation on the BEV feature map $z_j=g(P_j)$ to extract a corresponding feature vector, which is fed into the motion head, which is a Multi-Layer Perceptron (MLP), to predict the parameters of the SWGMM $p_\mathrm{swgmm}$.
\begin{equation}
p_\mathrm{swgmm}(\textbf{v}) = \sum_{k=1}^K w_k \mathcal{N}(\mathbf{v}|\boldsymbol{\mu}_k, \Sigma_k)    
\end{equation}
\begin{equation}
f_\mathrm{motion}(x, z) = [w_k(x, z), \boldsymbol{\mu}_k(x, z), \Sigma_k(x, z)]_{k=1}^K,    
\end{equation}
which is $K$ sets of Semi-Wrapped Normal Distribution $\mathcal{N}$ with weight $w_k$, mean $\boldsymbol{\mu}_k$ and covariance $\Sigma_k$. 
This formulation enables the model to learn continuous, multimodal distributions of human motion.

In addition to the motion head, our framework introduces the frequency head to predict frequency score in $[0,1]$ as map jointly from the BEV feature $z_j$:
\begin{equation}
f_\mathrm{freq}(z) = \sigma(q(z)),  
\end{equation}
where $\sigma$ is a sigmoid function. This score represents the spatial preference for where humans tend to exist or walk, complementary to the normalized motion distribution at each position.

Our architecture differs from NeMo-map in three aspects: First, while NeMo-map optimizes a 2D feature grid as learnable parameters, we replace this with a BEV feature map extracted from LiDAR points. This allows the model to relate knowledge across diverse geometric layouts and interpolate unobserved positions. Second, we intentionally exclude direct coordinate temporal dependencies that cannot be inferred solely from geometry. 
Third, our model is designed to jointly predict both the motion distribution and the spatial preference as a frequency score as explained above. The joint optimization with frequency scores encourages the model to learn naturally smooth motion patterns in flat regions as demonstrated in our experiments.

\subsection{Training process}
\subsubsection{Preprocess}
To prepare data for training, we first aggregate motion samples into a single coordinate system for each sequential record, utilizing the pose $T_j^c$ to cancel ego-motion, and then extract samples within a predefined spatial region, generating $N_j$ target motions for input $P_j$. This allows the model to access motion samples at timestamps decoupled from the specific LiDAR scans, thereby preventing the network from over-focusing on instantaneous pedestrian points instead of the underlying environmental geometry\footnote{Note that as the dataset scales in the future, this processing can be modified to accurately capture temporal changes within a sequence, such as obstacle positions.}.

The target frequency score $s$ is generated as follows: the annotated pedestrian bounding boxes are accumulated to a 2D histogram with a $0.1$ m resolution, and the counts are then divided by the standard deviation within each sequence. To remove peak variance unrelated to the true underlying flow frequency, we clip the count scores above $1$ to bound the final values within $[0, 1]$. The resulting score map is loaded as a local crop according to $T_j^c$ for each $P_j$.

\subsubsection{Grid Sampling}
Trajectory datasets contain many slow or stationary pedestrians, especially in indoor scenes, who generate redundant motion samples at identical spatial coordinates. Under a naive uniform sampling scheme across the aggregated samples, the model would relatively ignore dynamic samples and overfit to the spatial biases.

To mitigate this problem, we propose a sampling strategy, Grid sampling (GS). We group accumulated motion data into discrete spatial grid cells and select at most one motion sample per cell during training, thereby enforcing positional balance within $P_j$.

\subsubsection{Loss functions}
The motion head is trained using negative log-likelihood loss for SWGMM, as in NeMo-map.
\begin{equation}
L_\mathrm{motion} = - \frac{1}{N_j}\sum_{i=1}^{N_j}\log ( p_\mathrm{swgmm}(\textbf{v}_i|f_\mathrm{motion}) )
\end{equation}

We can use binary cross-entropy to learn the frequency scores $s$:
\begin{equation}
l(q, s) = - s\log( \sigma(q)) - (1-s) \log(1-\sigma(q)) 
\end{equation}

However, the frequency score label $s$ built by sparse data shows a ``positive-unlabeled" nature, and naive supervision may erroneously tell unobserved areas as zero frequency (``negative"), even in regions where humans likely exist (``positive") in long observation. 

To address this problem, we introduce two visibility mask-based modifications, ``PU" and ``PNi", for the frequency loss. We generate a visibility mask $m$ by calculating the maximum visible range in all tilt levels at each azimuth beam angle for LiDAR data. Invisible areas $m=0$ (black area in an example in \cref{fig:overview}) beyond the range are regarded as probably occupied and negative regions, since areas behind walls or inside large obstacles are always invisible.
PU strategy is to apply the unbiased form~\cite{NIPS2014pu} for Positive-unlabeled learning by approximating the positive sample ratio as the visible area ratio $r_j$. PNi strategy is to filter visible and unlabeled positions in the loss calculation, thereby preventing the model from learning potentially positive areas as negative.
The PU loss and the PNi loss for sample $j$ are written as follows:
\begin{equation}
L_{\mathrm{freq}}^{\mathrm{PU}}=
\frac{r_j}{|D_j^{P}|}\!\sum_{i \in D^P_j}\!\{l(q_i, s_i) \mathalpha{-} l(q_i, 0)\}
+ \frac{1}{|D_j^{U}|}\!\sum_{i \in D^{U}_j} l(q_i, 0),
\end{equation}
\begin{equation}
L_\mathrm{freq}^\mathrm{PNi} = \frac{1}{|D_j^{P}|+|D_j^{Ni}|}\{\sum_{i \in D^P_j} l(q_i, s_i)+\sum_{i \in D^{Ni}_j } l(q_i, 0)\},
\end{equation}
where $D_j^{P}$, $D_j^{U}$ and $D_j^{Ni}$ refers to the set of positions for $s_i>0$, $s_i=0$ and $(s_i=0) \& (m=0)$, respectively.
We experimentally evaluate these losses, enabling the model to learn robust motion patterns even when the training data is sparse.

\subsection{Motion Prediction Algorithm}
For downstream trajectory prediction, we propose a stable and deterministic algorithm by using mode velocity and reweighting the estimated prior.
Existing algorithm in CLiFF-LHMP~\cite{zhu2023clifflhmp} iteratively steps the position, samples motion at the current position, and updates the velocity angle. However, reliance on stochastic motion sampling leads to high variance across runs and jerky trajectories due to independent sampling at each time step.

Our deterministic algorithm is shown in ~\cref{alg:nemo_hmp}, where $P$ is LiDAR points and $\beta$ is a velocity update ratio. By interpreting the queried SWGMM $p_{M}$ as a motion prior, the algorithm computes a posterior distribution that penalizes large velocity updates by reweighting the prior with a Gaussian centered at the current velocity. Then, we choose the closest mode velocity $\textbf{v}_{M,t}$ from SWGMM with finite weight, starting from every mode at the first time step. 
The velocity angle is updated by mixing that of $\textbf{v}_{M,t}$ and the current one with the weight kernel $exp(-\beta(\cdot))$ for angle difference. Please refer to the Appendix for the full equations for each step. 

\begin{algorithm}[tb]
\caption{deterministic EgoNeMo-HMP}\label{alg:nemo_hmp}
\begin{algorithmic}[1]
\Require $T_{\text{obs}}$, $\textbf{x}_{t_0}$, $P$, $\Delta t$, $\beta$
\State $T_{\text{fut}} \gets \emptyset$
\State $z \gets g(P)$
\State $\textbf{v}_{\text{obs}} \gets \text{getObservedVelocity}(T_{\text{obs}})$
\State $s_{t_0} \gets (\textbf{x}_{t_0}, \textbf{v}_{\text{obs}})$ 
\For {$t = t_0 + 1, \dots, t_0 + T_p$}
    \State $\textbf{x}_t \gets \text{getNewPosition}(s_{t-1}, \Delta t)$
    \State $p_{M} \gets f_\mathrm{motion}(\textbf{x}_t, z)$
    \State $p_M \gets \text{reweightDistribution}(p_{M}, \textbf{v}_{t-1})$
    \State $\textbf{v}_{M,t} \gets \text{getVelocityMode}(p_M, \textbf{v}_{M,t-1})$
    \State $\textbf{v}_t \gets \text{predictVelocity}(\textbf{v}_{M,t}, \textbf{v}_{\text{obs}}; \beta)$
    \State $s_t \gets (\textbf{x}_t, \textbf{v}_t)$
    \State $T_{\text{fut}} \gets T_{\text{fut}} \cup \{s_t\}$
    \State $\textbf{v}_{\text{obs}} \gets \text{getObservedVelocity}(T_{\text{obs}} \cup T_{\text{fut}})$
\EndFor \\
\Return $T_{\text{fut}}$
\end{algorithmic}
\end{algorithm}
\section{Experiments}\label{sec:exp}
In this section, we present experiments to validate the efficacy of our method as MoD or trajectory prediction.

\subsection{Settings}
Experimental settings are explained below. Please refer to Appendix.~B for more details.

\noindent \textbf{Datasets:} Our experiments are mainly conducted on JRDB~\cite{martin2021jrdb, saadatnejad2023jrdb}, which provides pedestrian trajectories with LiDAR data from a robot's perspective, enabling us to train and evaluate using paired motion and LiDAR geometry across multiple indoor and outdoor locations.
We use the official train split for training, keeping $5$ sequences for a sanity check during training. The test set contains $10$ known- and $17$ unknown-location sequences.
As unknown test locations with larger difference, we also use the SiT dataset~\cite{bae2023sit}, which include public pedestrian roads or crosswalks.

\noindent \textbf{Evaluation metrics:} We report negative log-likelihoods (NLLs) and prediction errors for quantitative evaluation.
Since there is limited geometric evidence near the boundary of of the $20\,\mathrm{m} \times 20\,\mathrm{m}$ crop, we evaluate NLL within the central $10\,\mathrm{m} \times 10\,\mathrm{m}$ region (The evaluation range dependency is shown in Appendix~C). 
To reduce spatial bias in test samples, we do not simply average NLL over all motion samples. Instead, we compute the mean NLL within each $0.5\,\mathrm{m}$ grid cell and then average these cell-wise means. We also report NLL with a fixed angular variance of $20^\circ$ to assess the quality of the predicted motion modes. In the trajectory prediction experiments, we report Average Displacement Error (ADE) and Final Displacement Error (FDE) over the future $T$ steps, given past observations of up to $7$ steps at $3\,\mathrm{fps}$.
Although JRDB provides an official trajectory-prediction benchmark and protocol~\cite{saadatnejad2023jrdb}, we do not adopt it because its protocol would confound the evaluation of motion estimation and perception errors.

\noindent \textbf{Baselines:} The most relevant existing methods are BFF~\cite{francesco2024bff} and NeMo-map~\cite{zhu2026nemomap}. NeMo-map is the state-of-the-art site-specific MoD. BFF is a transferable model based on 2D occupancy maps (which we generate from LiDAR points, excluding the ground and ceiling). We extended it to BFF* with SWGMM outputs to ensure a fair comparison with continuous MoD approaches.  
A baseline of our framework $\textit{Motion}$ or $\textit{Ours-base}$ is a natural extension of NeMo-map to map BEV features from LiDAR points, instead of site-specific feature grids, to motion distributions. The ablation configurations include joint frequency head optimization ($\textit{Motion-Freq}$), that with grid sampling strategy ($\textit{Motion(GS)-Freq}$), and that with PU and PNi modifications in the frequency loss ($\textit{Motion(GS)-FreqPU}$ and $\textit{Motion(GS)-FreqPNi}$). \textit{Ours} shows the result for \textit{Motion(GS)-FreqPNi}. 

{\setlength{\tabcolsep}{4pt}
\begin{table}
    \centering
        \caption{NLL comparison with existing MoD methods. JRDB-Known* refer to three sequences with clear spatial overlaps with train sequences, gates-ai-lab-2019-04-17\_0, huang-2-2019-01-25\_1, and tressider-2019-03-16\_2 from known location sequences.}
    \label{tab:mod_bench}
    \begin{tabular}{lccc}
    \toprule
Model & JRDB-Known* & JRDB-Unknown & SiT \\
\midrule
NeMo & 12.0 (1.16) & N/A & N/A \\
BFF* & \textbf{1.01} (0.06) & 1.53 (0.05) & 1.34 (0.12) \\
Ours & 1.29 (0.16) & \textbf{1.42} (0.05) & \textbf{1.24} (0.11) \\
\bottomrule
    \end{tabular}
\end{table}
}

\begin{table*}[th]
\centering
    \caption{NLL comparison in JRDB and SiT dataset. \textit{Motion} corresponds to the baseline that simply extends NeMo-map~\cite{zhu2026nemomap}. Each value represents the mean of three trials, with the standard deviation shown in parentheses. The lowest mean value per column is in bold.}
    \label{tab:nll_jrdb} 
\begin{tabular}{lcccccc}
\toprule
 & \multicolumn{2}{c}{JRDB-Known} & \multicolumn{2}{c}{JRDB-Unknown}& \multicolumn{2}{c}{SiT} \\
Method & NLL & ($20^\circ$ var) & NLL & ($20^\circ$ var) & NLL & ($20^\circ$ var) \\
\midrule
Motion & 1.22 (0.08) & 4.12 (0.61) & 1.79 (0.29) & 4.66 (0.35) & 1.63 (0.04) & 4.68 (0.86) \\
Motion-Freq & 1.14 (0.03) & 3.68 (0.12) & 1.67 (0.09) & 4.03 (0.39) & 1.45 (0.07) & 3.82 (0.74) \\
Motion(GS)-Freq & 1.14 (0.01) & 2.82 (0.55) & 1.47 (0.08) & 3.34 (0.48) & \textbf{1.22} (0.10) & \textbf{2.80} (0.46) \\
Motion(GS)-FreqPU &\textbf{ 0.95} (0.03) & 3.08 (0.69) & \textbf{1.33} (0.03) & 3.88 (0.66) & 1.22 (0.01) & 3.43 (1.35) \\
Motion(GS)-FreqPNi & 1.12 (0.07) & \textbf{2.54} (0.27) & 1.42 (0.05) & \textbf{2.98} (0.51) & 1.24 (0.11) & 2.82 (0.74) \\
\bottomrule
\end{tabular}
\end{table*}

\subsection{Results}
\subsubsection{Benchmarks} 
Comparative evaluations against existing methods are summarized in \cref{tab:mod_bench}. 
While NeMo-map struggles due to data sparsity per location, transferable models leveraging multi-location training (BFF* and Ours) achieve substantially lower NLL values. Crucially, our method achieves superior performance in JRDB-Unknown and SiT, validating its enhanced zero-shot generalization capability to unknown locations. We refer the reader to Appendix.~D for a detailed evaluation.

\subsubsection{Ablation study}
Ablation results are shown in ~\cref{tab:nll_jrdb}. Performance on the known locations of the JRDB dataset can be interpreted as the upper bound for unknown locations. The known-location results are comparable across all variants, and unknown locations naturally exhibit higher NLLs. For unknown locations, the baseline $\textit{Motion}$ model experiences severe NLL degradation. In contrast, $\textit{Motion(GS)-Freq}$ effectively suppresses this degradation, indicating more robust generalization. Notably, $\textit{Motion(GS)-FreqPU}$ achieved the minimum NLL when computed directly from the model outputs. However, this model suffers from over-inflated angular variance, which minimizes the NLL loss but fails to capture the distinct physical modes of pedestrian flow directions. 
This becomes evident when the predicted angular variance is replaced with a fixed value: the PU-based loss delivers the worst performance among the variants with GS, whereas $\textit{Motion(GS)-FreqPNi}$ achieves the best average NLL. $\textit{Motion(GS)-Freq}$ or $\textit{Motion(GS)-FreqPNi}$ show comparable NLLs on JRDB-Unknown and SiT, demonstrating zero-shot transferability of our method.

\begin{figure*}[tbp]
    \centering
    \begin{tabular}{cccc}
    Reference & Input & Motion & Motion-Freq \\
    \includegraphics[width=0.45\linewidth, trim=10 10 450 10, clip]{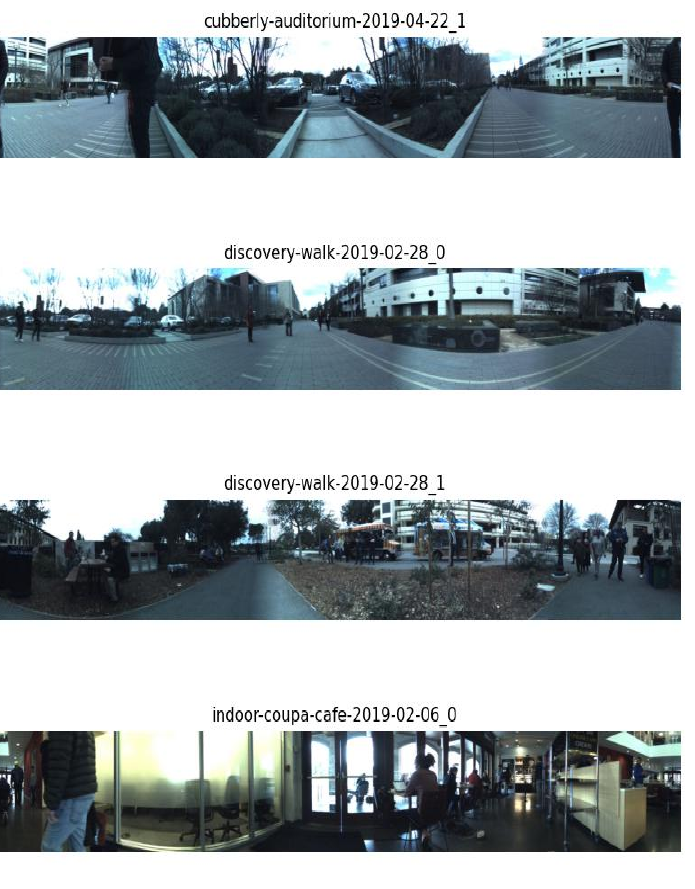} &
    \includegraphics[width=0.15\linewidth, trim=0 270 0 0, clip]{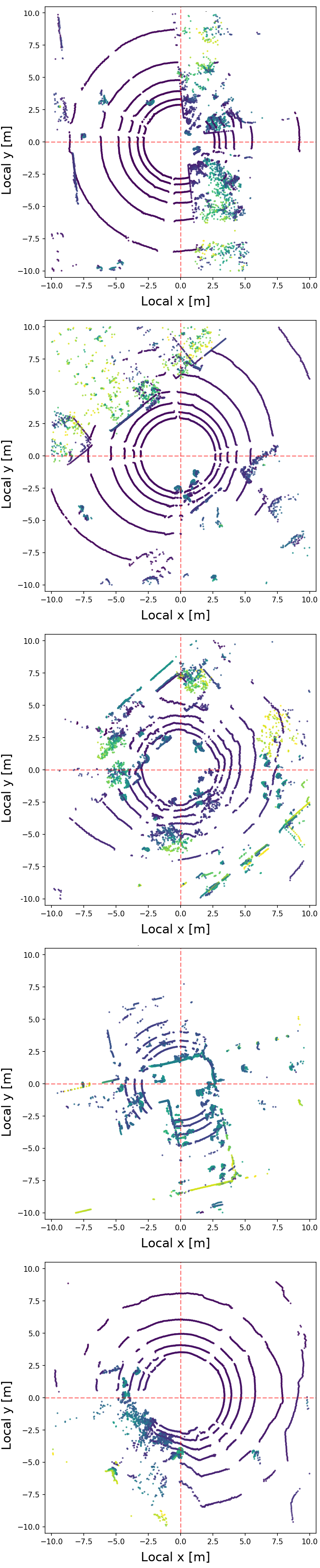} &
    \includegraphics[width=0.15\linewidth, trim=890 880 0 0, clip]{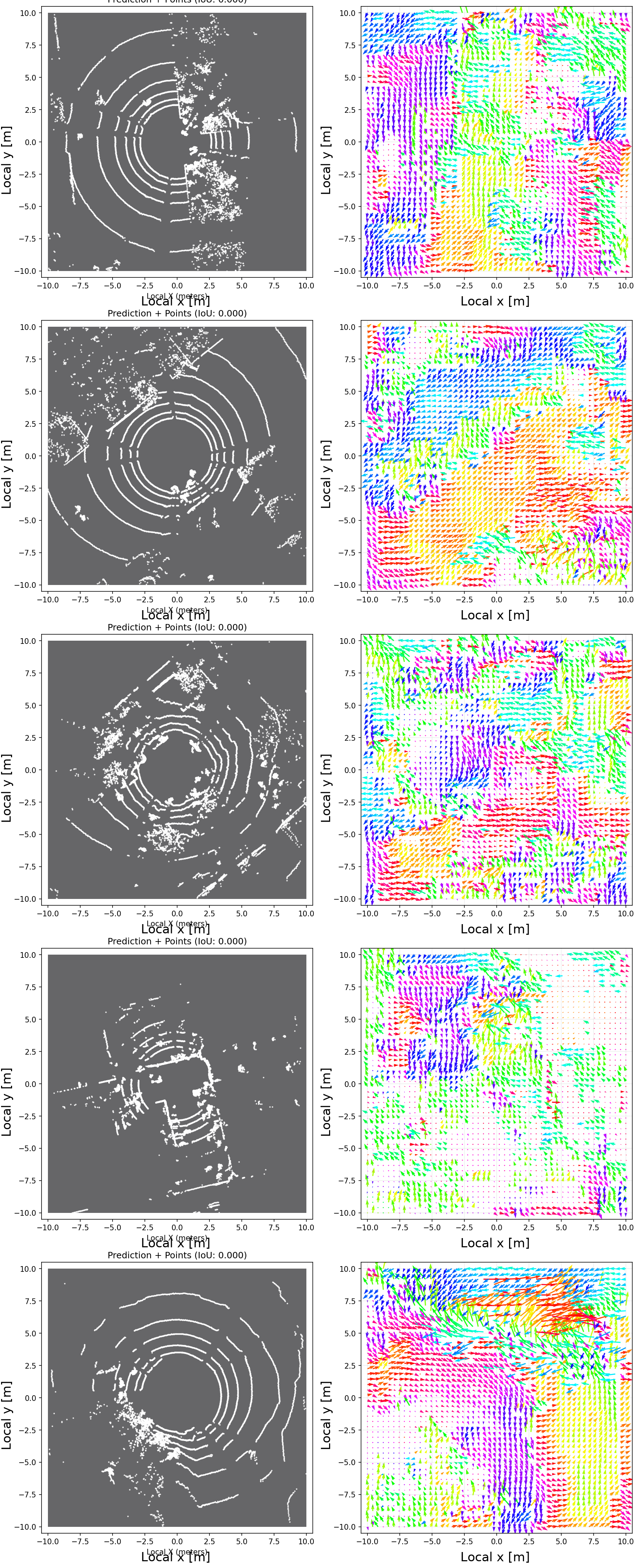} &
    \includegraphics[width=0.15\linewidth, trim=890 880 0 0, clip]{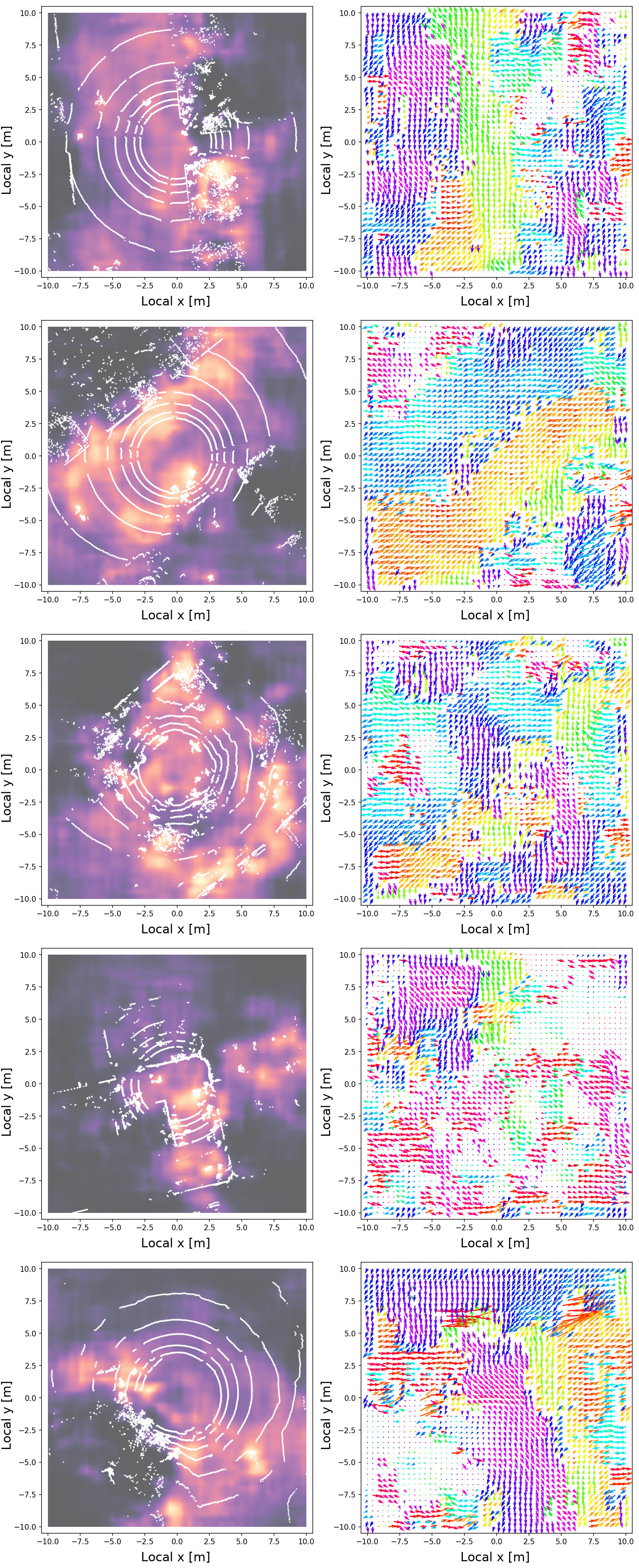} 
    \end{tabular}
    \caption{Outputs of \textit{Motion} and \textit{Motion-Freq} on JRDB unknown locations. In the motion maps, arrows visualize the velocity modes; opacity represents weight, arrow length represents speed, and hue represents direction using a cyclic color mapping. The same hue corresponds to the same direction across all maps. The reference column shows the sequence names and stitched images, horizontally compressed for visualization. The input column shows LiDAR data as BEV with height-based color from navy to yellow.
}
    \label{fig:motion-outputs}
\end{figure*}

~\Cref{fig:motion-outputs} summarizes a qualitative comparison of motion outputs, queried at a $100 \times 100$ grid, with the reference images and LiDAR inputs.
In the motion maps, arrows visualize the velocity modes; arrow direction and length represent heading and speed, respectively, while arrow hue encodes heading angle using a cyclic color mapping: rightward, upward, leftward, and downward motions correspond to red, light green, cyan, and purple, respectively.
The LiDAR plots show points in the BEV space with height-based color from navy to yellow, and the navy circle patterns appear on flat ground around the robot at the origin. The major direction looks similar between \textit{Motion} and \textit{Motion-Freq}, but \textit{Motion-Freq} has a smoother map along flat ground roads, thanks to the frequency score learning. The predicted motion shows flow along the ground shape with right-handed tendency in the top two rows, and exhibits almost zero velocity (shown as short arrows-like points) around cafe tables in the bottom row, along our intuition.

\begin{figure*}[tbp]
    \centering
    \begin{tabular}{c@{\hspace{5pt}}c@{\hspace{5pt}}c}
    Motion(GS)-Freq & Motion(GS)-FreqPU & Motion(GS)-FreqPNi \\
    \includegraphics[width=0.31\linewidth, trim=0 2640 0 10, clip]{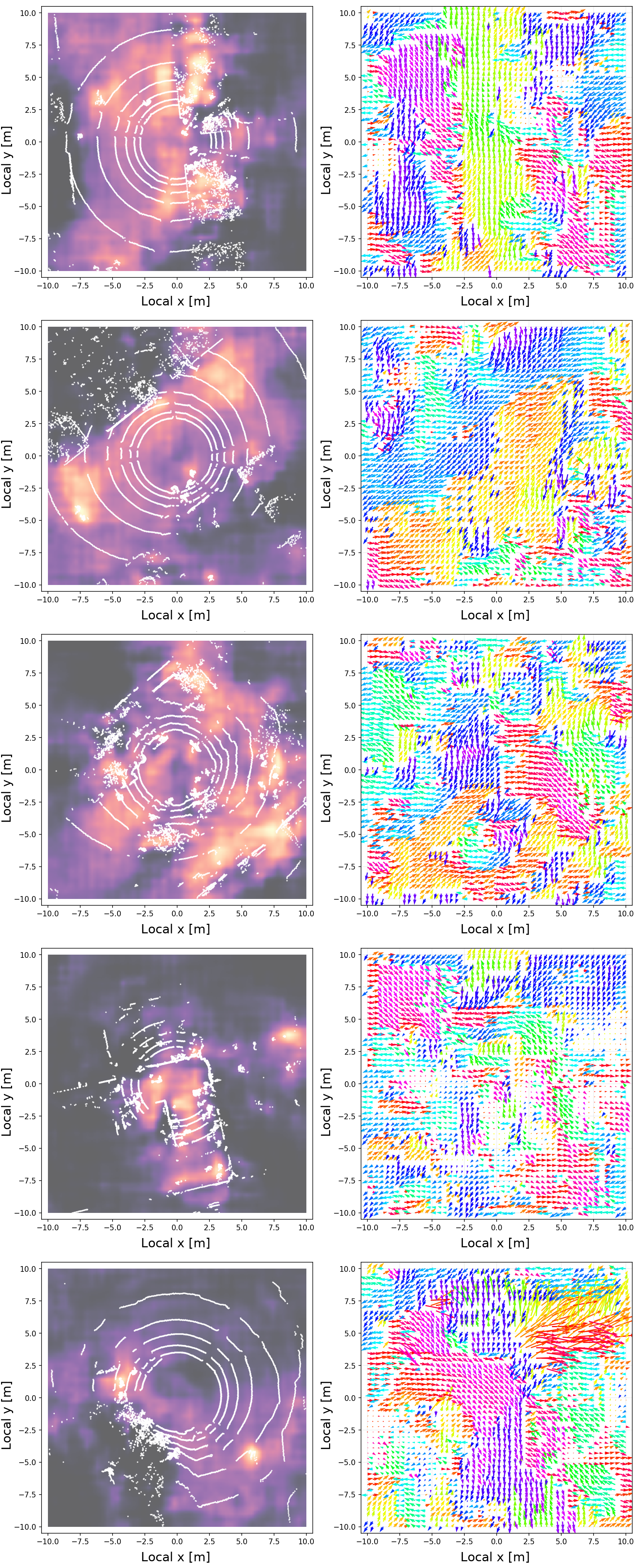} &
    
    \includegraphics[width=0.31\linewidth, trim=0 2640 0 10, clip]{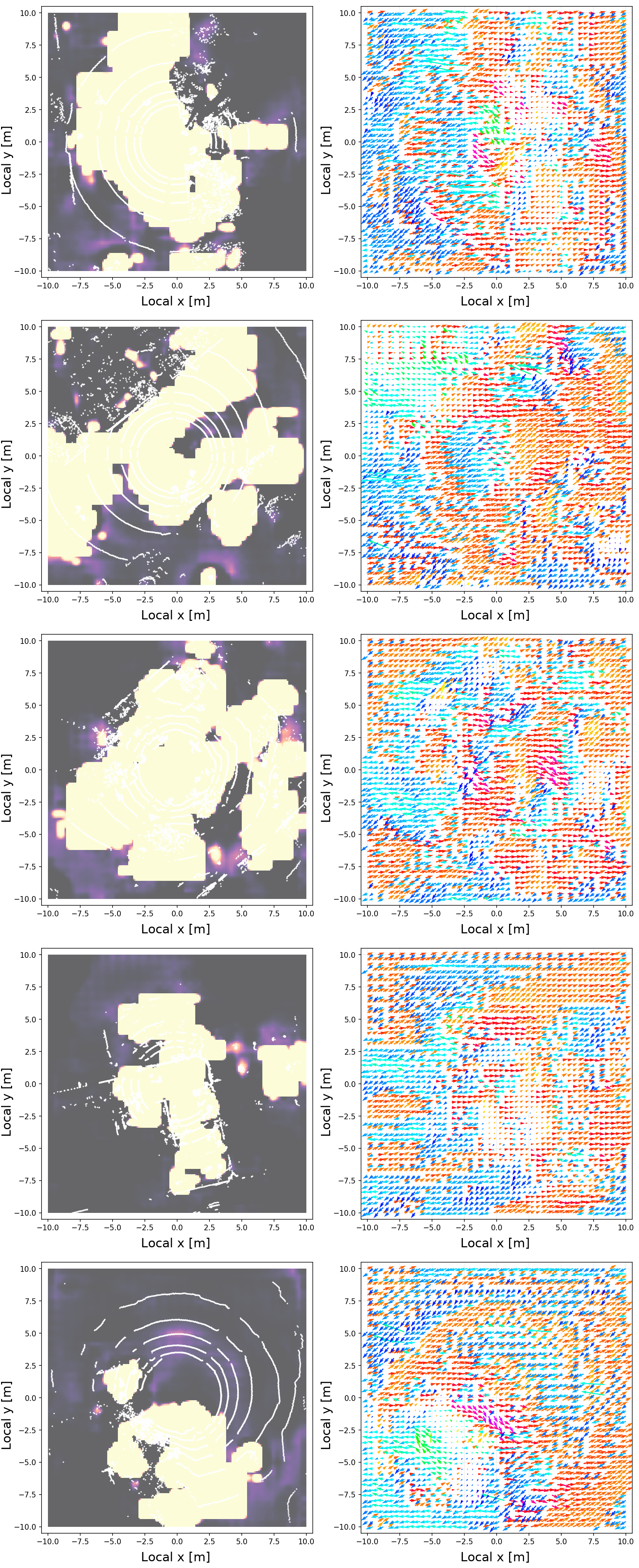} &

    \includegraphics[width=0.31\linewidth, trim=0 2640 0 10, clip]{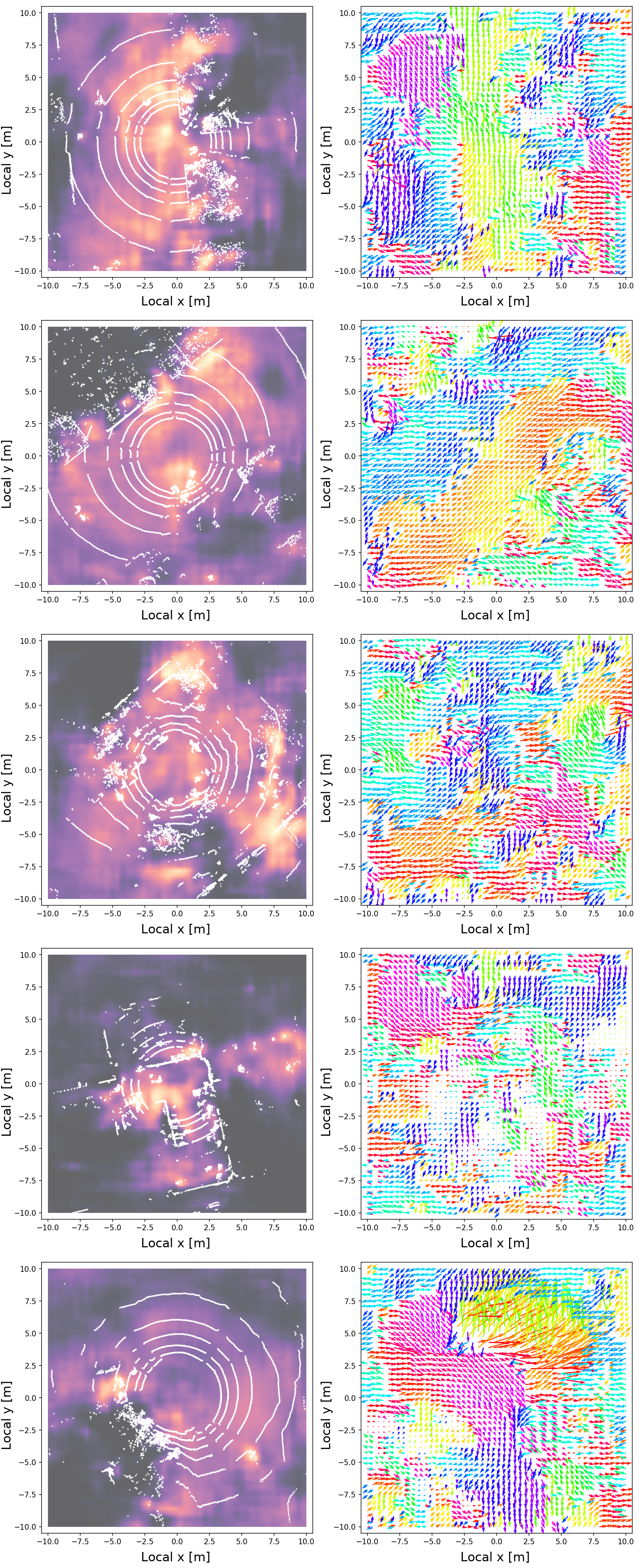}
    \end{tabular}
    \vspace{-10pt}
    \caption{Frequency score and motion outputs of \textit{Motion(GS)-Freq}, \textit{Motion(GS)-FreqPU}, and \textit{Motion(GS)-FreqPNi}. For each method, the left panel shows the predicted frequency score (brighter color means higher score), and the right panel shows the motion map. Samples and arrow encoding of motion maps are the top two rows of \cref{fig:motion-outputs}, the rest are in the Appendix Fig.~4.
    }
    \label{fig:freq_motion_outputs}
\end{figure*}

~\Cref{fig:freq_motion_outputs} summarizes a qualitative comparison of three frequency loss variants.
Although these variants differ only in the frequency-map training loss, they also produce clear differences in the motion outputs. \textit{Motion(GS)-FreqPU} shows blocky frequency maps due to the characteristics of the PU loss nature and shows less variation in mean angles, which is consistent with its degraded fixed-variance NLL. 
$\textit{Motion(GS)-FreqPNi}$ yields higher frequency scores within the visible range than $\textit{Motion(GS)-Freq}$ by excluding visible yet unlabeled areas from the loss calculation.

To evaluate the robustness to train data scarcity, we train models with fewer locations (only locations in the known test set) and shorter durations (the first half of each sequence). While a reduction in training data causes a consistent increase in NLL compared to ~\cref{tab:nll_jrdb}, the baseline $\textit{Motion}$ exhibits a larger degradation for shorter durations. The three variants featuring both $\textit{Freq}$ and $\textit{GS}$ components exhibit less degradation, demonstrating the robustness to short-duration biases as intended. Specifically for fewer locations, \textit{Motion(GS)-FreqPU} keeps low NLL even with fixed variance unlike other experiments, which indicates another robustness to the geometric variation problem. 

\begin{table*}[tb]
\centering
\setlength{\tabcolsep}{5pt}
\caption{NLL comparison in JRDB dataset for reduced train data. Each value represents the mean of three trials, with the standard deviation shown in parentheses. The lowest mean value in each column is highlighted in bold. The method column shows short-hand names (M, F, and G) instead of Motion, Freq, and GS, respectively.}
\label{tab:nll_jrdb_small}
\begin{tabular}{lcccccccc}
    \toprule
 & \multicolumn{4}{c}{Fewer locations} & \multicolumn{4}{c}{Shorter durations} \\
 & \multicolumn{2}{c}{JRDB-Known} & \multicolumn{2}{c}{JRDB-Unknown} & \multicolumn{2}{c}{JRDB-Known} & \multicolumn{2}{c}{JRDB-Unknown} \\
Method & NLL & ($20^\circ$ var) & NLL & ($20^\circ$ var) & NLL & ($20^\circ$ var) & NLL & ($20^\circ$ var) \\
\midrule
M only & 1.24 (0.08) & 4.70 (0.54) & 1.81 (0.09) & 5.09 (0.16)  & 1.50 (0.14) & 4.37 (0.58) & 2.26 (0.54) & 5.31 (0.74) \\
M-F & 1.23 (0.06) & 4.20 (0.48) & 1.81 (0.03) & 5.40 (0.54) & 1.43 (0.10) & 4.34 (0.23) & 1.96 (0.05) & 5.17 (0.43) \\
M(G)-F & 1.51 (0.01) & 3.40 (0.43) & 1.87 (0.05) & 4.11 (0.34)  & 1.30 (0.02) & \textbf{2.89} (0.14) & 1.75 (0.06) & \textbf{3.78} (0.05) \\
M(G)-FPU & \textbf{0.97} (0.04) & \textbf{3.22} (0.45) & \textbf{1.40} (0.02) & \textbf{4.03} (0.64) & \textbf{1.01} (0.03) & 3.20 (0.35) & \textbf{1.37} (0.06) & 4.22 (0.76)\\
M(G)-FPNi & 1.51 (0.03) & 3.36 (0.40) & 1.91 (0.03) & 4.17 (0.46) & 1.31 (0.01) & 3.01 (0.23) & 1.73 (0.01) & 3.91 (0.19) \\
\bottomrule
\end{tabular}
\end{table*}

\subsubsection{Trajectory Prediction Application} 
To demonstrate the practicality of our MoD and prediction algorithm, we performed trajectory prediction evaluation. We compare our method with MoD-based \textit{NeMo}, the stochastic sampling method~\cite{zhu2023clifflhmp,zhu2026nemomap}\footnote{\textit{NeMo} is evaluated with additional averaging over all possible $K$ output selections from $20$ randomly sampled trajectories.} on our MoD, and recent non-MoD predictors ST~\cite{saadatnejad2024socialtransmotion} and ViTE~\cite{li2026vite}. 
Non-MoD models can be applied to unseen locations by ignoring environmental conditions, focusing on trajectories and inter-agent interactions. 
Note that MoD-based predictors focus on environmental conditions, ignoring inter-agent interactions.

Our deterministic algorithm achieves lower error than \textit{NeMo}. Also, we can see the MoD improvement from \textit{Ours-base} reduces prediction error not only NLL. Our method does not surpass the non-MoD predictor ViTE (results with longer $T$s is shown in Appendix Tab.~6). 
Integrating our transferable MoD as a complementary information to interaction-aware predictors such as ViTE is left for future work.

\begin{table}
    \centering
    \setlength{\tabcolsep}{5pt}
        \caption{
        Trajectory prediction errors on JRDB. $K$ is the number of modes. \textit{NeMo} uses CLiFF-LHMP algorithm~\cite{zhu2023clifflhmp,zhu2026nemomap} on the same MoD as \textit{Ours}. All results are averaged over 3 training runs.
        }
    \label{tab:pred_exp}
    \begin{tabular}{l@{\hspace{3pt}}c|cc|cc|cc}
    \toprule
 & & \multicolumn{2}{c}{T=12} & \multicolumn{2}{c}{T=18} & \multicolumn{2}{c}{T=24} \\
Method & K & ADE & FDE & ADE & FDE & ADE & FDE \\
\midrule
ST~\cite{saadatnejad2024socialtransmotion} & 1 & 0.54 & 0.99 & 0.79 & 1.57 & 1.05 & 2.14 \\
ViTE~\cite{li2026vite} & 3 & \textbf{0.36} & \textbf{0.57} & \textbf{0.52} & \textbf{0.84} & \textbf{0.67} & \textbf{1.11} \\
\midrule
NeMo~\cite{zhu2026nemomap} & 1 & 0.44 & 0.82 & 0.63 & 1.23 & 0.81 & 1.60 \\
NeMo~\cite{zhu2026nemomap} & 3 & 0.42 & 0.80 & 0.62 & 1.20 & 0.79 & 1.57 \\
Ours-base & 3 & 0.42 & 0.79 & 0.61 & 1.19 & 0.78 & 1.55 \\
Ours & 3 & \textbf{0.41} & \textbf{0.78} & \textbf{0.60} & \textbf{1.16} & \textbf{0.77} & \textbf{1.51} \\
\bottomrule
    \end{tabular}
\end{table}

We report the inference speed measured on a Quadro RTX 5000. The MoD estimation runs at $4.0$ ms to evaluate at $100$ positions per LiDAR scan. Our trajectory prediction runs at around $23.6$ ms for $K=3$ modes and $T=12$. Both are fast enough for robot operation, such as navigation. 

\subsection{Limitations}
Our model relies on egocentric LiDAR points in $20\,\mathrm{m} \times 20\,\mathrm{m}$ crops, so it has limitations: (i) the output in invisible regions is unreliable because there are no geometric cues or enough trajectory data for training; (ii) it cannot consider far doors or walls, which leads to weakness in spacious outdoor scenes; (iii) site-specific time dependency in NeMo-map can not be predicted. There are degradations near the boundaries due to occlusions or insufficient geometry inputs, but our method is not sensitive to occlusions by nearby persons. See Appendix.~C for related experiments.


\section{Conclusion}
We proposed a transferable Map of Dynamics (MoD) framework that instantly generalizes to unknown environments using only egocentric 3D LiDAR points.
Extending recent advances in neural implicit modeling, our framework realized a continuous, LiDAR-based MoD estimator trained across diverse environments. To mitigate the inherent sparsity and temporal bias of real-world trajectory data, we introduced a position-balanced sampling strategy and a multi-task learning architecture that jointly predicts motion distributions and a frequency score map. The frequency component was further augmented with an visibility-aware frequency loss to mitigate incomplete frequency information. Comprehensive experiments showed that our method successfully reconstructs motion fields from highly limited train data. Our framework achieved immediate zero-shot transfer to unseen environments, enabling reliable pedestrian trajectory prediction from a single, instantaneous LiDAR scan. Promising directions for future work include incorporating RGB-based semantic guidance for more accurate estimation and integrating into trajectory predictors. 
\clearpage 
{
    \small
    \bibliographystyle{ieeenat_fullname}
    \bibliography{main}
}

\title{

}

\setcounter{page}{1} 
\setcounter{section}{0}
\setcounter{table}{0}
\setcounter{figure}{0}
\renewcommand{\thesection}{\Alph{section}}
\renewcommand{\thetable}{S\arabic{table}}
\renewcommand{\thefigure}{S\arabic{figure}}

\twocolumn[
  \begin{center}
  {\Large \bf Appendix of ``EgoNeMo: Transferable Map of Pedestrian Dynamics via Egocentric LiDAR Scan''}
    \vspace{1.5em}
  \end{center}
]

\section{Motion Prediction Algorithm Details}
This section provides a detailed description of the trajectory prediction algorithm (\cref{alg:nemo_hmp}) presented in the main paper. The primary modification over CLiFF-LHMP~\cite{zhu2023clifflhmp} is the direct utilization of the mode velocity rather than stochastic sampling. Furthermore, several key modifications are introduced to handle multimodal distributions.

\noindent \textbf{Notation:} 
Let $\mathbf{P}$ denote the input point clouds, and $T_{\text{obs}}$ represent the past observed pedestrian trajectory containing the 2D position $\mathbf{x} \in \mathbb{R}^2$ and velocity $\mathbf{v}$ per frame. The velocity is parameterized as $\mathbf{v} = (\rho \cos\theta, \rho \sin\theta)$, where $\rho \in \mathbb{R}^+$ denotes the speed and $\theta \in [0, 2\pi)$ represents the angle. The state vector at any given time step is defined as $s = (\mathbf{x}, \mathbf{v})$. 

An encoder $g$ maps the point cloud $\mathbf{P}$ to a Bird's-Eye-View (BEV) feature map $z$. A motion network $f_{\text{motion}}$ then maps the feature $z$ queried at an arbitrary position $\mathbf{x}_i$ to a Semi-Wrapped Gaussian Mixture Model (SWGMM)~\cite{Roy16SWGMM}. The parameters governing the algorithm include the time step size $\Delta t$, the blend parameter $\beta$, and the observation window scale $\sigma_t$.

The core procedures, alongside our specific modifications, are detailed below. The parameter values utilized throughout this study are summarized in \cref{tab:param}.

\subsection{Common Steps}
Initially, the BEV feature map $z$ is extracted from $\mathbf{P}$, and the function $\text{getObservedVelocity}$ computes the current velocity from $T_{\text{obs}}$. This initial velocity is computed via a weighted average process over historical frames with an exponential decay rate of $\exp(-\Delta t/\sigma_t)$ per time step.
The remaining steps iteratively project the pedestrian's state by updating the position via $\text{getNewPosition}$:
\begin{equation}
\mathbf{x}_t = \mathbf{x}_{t-1} + \mathbf{v}_{t-1} \Delta t,
\end{equation}
followed by a dynamic update of the current velocity.
 
\subsection{Mode Reweighting (\texttt{reweightDistribution})}
By interpreting the queried SWGMM as a continuous motion prior, we compute a posterior distribution that penalizes physically unrealistic or abrupt velocity updates. This is achieved by multiplying the prior distribution by a Gaussian transition probability centered at the current velocity $\mathbf{v}_{t-1}$. We conducted parameter searches to determine the speed variance $\sigma_{\rho}^2$ and angle variance $\sigma_{\theta}^2$ using NeMo-map~\cite{zhu2026nemomap} evaluated on the ATC dataset~\cite{brscic2013person}.

\subsection{Mode Extraction (\texttt{getVelocityMode})}
Since the SWGMM-based distribution has multiple modes, we dynamically select the closest mode whose weight exceeds a small threshold $w_{\text{th}}$ to ensure temporal consistency. Starting from each mode at the initial time step $t_0$, the algorithm rolls out $K$ deterministic trajectories (where $K=3$ in our evaluation). Compared to the original CLiFF-LHMP which relies on stochastic sampling at every individual time step, this formulation substantially reduces the number of sample trajectories required to achieve statistically stable forecasting results.

\subsection{Velocity Update (\texttt{predictVelocity})}
The pedestrian velocity is updated by blending the angular component in a manner similar to CLiFF-map. However, our formulation anchors the adjustment to the observed angle $\theta_{\text{obs}}$ instead of the previous step's heading $\theta_{t-1}$:
\begin{equation}
\theta_t = \theta_{t-1} + k(\theta_{M,t}-\theta_{\text{obs}}) \cdot (\theta_{M,t}-\theta_{\text{obs}}),
\end{equation}
where $\theta_{\text{obs}}$ is iteratively updated via $\text{getObservedVelocity}$. The blending kernel function $k$ is defined as:
\begin{equation}
k(\theta_{M,t}-\theta_{\text{obs}}) = \exp\left(-\beta (\theta_{M,t}-\theta_{\text{obs}})^2\right).
\end{equation}
This anchoring strategy mitigates cyclic trajectory artifacts that often arise from consistent mode extractions.

\begin{algorithm}[tb]
\caption{Deterministic EgoNeMo-HMP}\label{alg:nemo_hmp}
\begin{algorithmic}[1]
\Require $T_{\text{obs}}$, $\mathbf{x}_{t_0}$, $\mathbf{P}$, $\Delta t$, $\beta$, $\sigma_t$
\State $T_{\text{fut}} \gets \emptyset$
\State $z \gets g(\mathbf{P})$
\State $\mathbf{v}_{\text{obs}} \gets \text{getObservedVelocity}(T_{\text{obs}}; \sigma_t)$
\State $s_{t_0} \gets (\mathbf{x}_{t_0}, \mathbf{v}_{\text{obs}})$ 
\For {$t = t_0 + 1, \dots, t_0 + T_p$}
    \State $\mathbf{x}_t \gets \text{getNewPosition}(s_{t-1}, \Delta t)$
    \State $p_{M} \gets f_\mathrm{motion}(\mathbf{x}_t, z)$
    \State $p_M \gets \text{reweightDistribution}(p_{M}, \mathbf{v}_{t-1})$
    \State $\mathbf{v}_{M,t} \gets \text{getVelocityMode}(p_M, \mathbf{v}_{M,t-1})$
    \State $\mathbf{v}_t \gets \text{predictVelocity}(\mathbf{v}_{M,t}, \mathbf{v}_{\text{obs}}; \beta)$
    \State $s_t \gets (\mathbf{x}_t, \mathbf{v}_t)$
    \State $T_{\text{fut}} \gets T_{\text{fut}} \cup \{s_t\}$
    \State $\mathbf{v}_{\text{obs}} \gets \text{getObservedVelocity}(T_{\text{obs}} \cup T_{\text{fut}}; \sigma_t)$
\EndFor \\
\Return $T_{\text{fut}}$
\end{algorithmic}
\end{algorithm}


\begin{table}[ht]
\centering
\caption{Hyperparameter configurations utilized in the trajectory prediction evaluation.}
\label{tab:param}
\begin{tabular}{lc}
\toprule
Parameter & Value \\
\midrule
Time step size $\Delta t$ & $0.33$ [s] \\
Kernel time scale $\sigma_t$ & $1.5$ [s] \\
Mode weight threshold $w_{\text{th}}$ & $0.05$ \\
Velocity blend parameter $\beta$ & $5.0$ [$\text{rad}^{-2}$] \\
Prior speed variance $\sigma_{\rho}^2$ & $(0.01\text{ [m/s]})^2$ \\
Prior angle variance $\sigma_{\theta}^2$ & $(0.4\text{ [rad/s]})^2$ \\
\bottomrule
\end{tabular}
\end{table}

\section{Experimental details}\label{sec:exp_supplement}
Main dataset in our experiments is JRDB~\cite{martin2021jrdb, saadatnejad2023jrdb}, which provides pedestrian trajectories with LiDAR data from a robot's perspective. We use the official train split for training, keeping $5$ sequences for a sanity check during training. The test set contains $10$ known location sequences and $17$ unknown sequences.
We treated below sequences as JRDB-Known:
    'gates-ai-lab-2019-04-17\_0',
    'gates-to-clark-2019-02-28\_0',
    'huang-2-2019-01-25\_1',
    'nvidia-aud-2019-01-25\_0',
    'nvidia-aud-2019-04-18\_1',
    'nvidia-aud-2019-04-18\_2',
    'tressider-2019-03-16\_2',
    'tressider-2019-04-26\_0',
    'tressider-2019-04-26\_1',
    'tressider-2019-04-26\_3'.

We also used the annotated sequences in SiT dataset~\cite{bae2023sit} as unknown location data, which is a dataset collected using the same LiDAR sensor on a different robot. Note that we don't use this during training at all and this dataset provides test environments far from JRDB.

Our BEV encoder is a PointPillars~\cite{pointpillars2019}-based architecture with a pillar size of $0.2$ m $\times$ $0.2$ m, which outputs a $100 \times 100$ BEV feature map with $64$ channels. The motion head is an MLP with hidden sizes $[128, 64]$ and ReLU activations, and the number of modes is set to $3$. The frequency head is a convolutional neural network with the number of channels $[128,64,64]$ and kernel sizes $[3,3,1]$ and two batch normalization layers between layers. Every model is trained by AdamW~\cite{loshchilov2018decoupled} with learning rate $0.001$, batch size $8$, and weight decay $1\times10^{-5}$ for $100$ epochs, using LiDAR data every $20$ frames under random azimuth rotations.

In our trajectory prediction experiment, Social-transmotion~\cite{saadatnejad2024socialtransmotion} is trained by shuffling "ego agent" to predict trajectory and evaluated by running model per agent, differently from their official code.

\section{Analysis of Limitation}\label{sec:limit}

\subsection{Reliability at Boundary}
We partition the output Bird's-Eye-View (BEV) space into four non-overlapping, concentric square regions centered at the ego-sensor origin $(0, 0)$:
\begin{itemize}
    \item Zone 1: $[0.0\,\text{m}, 2.5\,\text{m}]$
    \item Zone 2: $(2.5\,\text{m}, 5.0\,\text{m}]$
    \item Zone 3: $(5.0\,\text{m}, 7.5\,\text{m}]$
    \item Zone 4: $(7.5\,\text{m}, 10.0\,\text{m}]$
\end{itemize}
Each interval defines the absolute distance boundary for both $x$ and $y$ axes.

\Cref{tab:boundary,tab:boundary_20deg} show NLLs and NLLs with fixed variance per zone respectively. Performance generally degrades toward the boundary.
{
\setlength{\tabcolsep}{5pt}
\begin{table*}
\centering
\caption{Spatial range analysis across regions from center to boundary. Negative Log-Likelihood (NLL) is evaluated across non-overlapping region bins centered at the sensor origin. Ranges indicate the lower and upper bounding thresholds for both $x$ and $y$ absolute coordinates in meters. Each value represents the mean of three trials, with the standard deviation shown in parentheses. The lowest mean value in each column is highlighted in bold.}
\label{tab:boundary}
\begin{tabular}{l|cccc|cccc}
\toprule
 & \multicolumn{4}{c}{JRDB-Known} & \multicolumn{4}{c}{JRDB-Unknown} \\
Method & [0.0, 2.5] & [2.5, 5.0] & [5.0, 7.5] & [7.5, 10.0] &  [0.0, 2.5] & [2.5, 5.0] & [5.0, 7.5] & [7.5, 10.0] \\
\midrule
M only & 1.16 (0.07) & 1.19 (0.09) & 1.53 (0.07) & 1.68 (0.02) & 1.71 (0.19) & 1.76 (0.30) & 1.98 (0.37) & 1.92 (0.17) \\
M-F & 1.11 (0.07) & 1.11 (0.03) & 1.44 (0.08) & 1.54 (0.04) & 1.61 (0.15) & 1.67 (0.08) & 1.83 (0.09) & 1.87 (0.10) \\
M(G)-F & 1.05 (0.04) & 1.12 (0.05) & 1.26 (0.03) & 1.39 (0.06) & 1.36 (0.04) & 1.47 (0.09) & 1.69 (0.13) & 1.78 (0.19) \\
M(G)-FPU & \textbf{0.87} (0.06) & \textbf{0.95} (0.03) & \textbf{1.08} (0.02) & \textbf{1.20} (0.02) & \textbf{1.28} (0.03) & \textbf{1.32} (0.04) & \textbf{1.47} (0.07) & \textbf{1.62} (0.10) \\
M(G)-FPNi & 1.06 (0.05) & 1.07 (0.08) & 1.29 (0.03) & 1.44 (0.02) & 1.43 (0.06) & 1.40 (0.05) & 1.61 (0.08) & 1.70 (0.07) \\
\bottomrule
    \end{tabular}
\end{table*}
}

{\setlength{\tabcolsep}{5pt}
\begin{table*}
\centering
\caption{Spatial range analysis across regions from center to boundary. Negative Log-Likelihood (NLL) with a fixed angular variance $20^\circ$ is evaluated across non-overlapping region bins centered at the sensor origin. Ranges indicate the lower and upper bounding thresholds for both $x$ and $y$ absolute coordinates in meters. Each value represents the mean of three trials, with the standard deviation shown in parentheses. The lowest mean value in each column is highlighted in bold.}
\label{tab:boundary_20deg}
\begin{tabular}{l|cccc|cccc}
\toprule
 & \multicolumn{4}{c}{JRDB-Known} & \multicolumn{4}{c}{JRDB-Unknown} \\
Method & [0.0, 2.5] & [2.5, 5.0] & [5.0, 7.5] & [7.5, 10.0] &  [0.0, 2.5] & [2.5, 5.0] & [5.0, 7.5] & [7.5, 10.0] \\
\midrule
M only & 3.93 (1.04) & 4.13 (0.57) & 4.92 (0.60) & 5.37 (0.51) & 4.61 (0.90) & 4.55 (0.32) & 5.30 (0.30) & 5.55 (0.67) \\
M-F & 3.35 (0.55) & 3.70 (0.10) & 4.54 (0.50) & 4.73 (0.52) & 3.97 (0.25) & 3.99 (0.32) & 4.46 (0.53) & 4.76 (0.24) \\
M(G)-F & 2.32 (0.42) & 2.87 (0.63) & 3.21 (0.57) & 3.53 (0.55) & \textbf{2.73} (0.28) & 3.43 (0.54) & 4.03 (0.58) & 4.25 (0.71) \\
M(G)-FPU & 2.84 (1.20) & 3.12 (0.66) & 3.33 (0.21) & 3.53 (0.15) & 3.51 (1.03) & 3.89 (0.61) & 4.27 (0.71) & 4.49 (0.40) \\
M(G)-FPNi & \textbf{2.23} (0.35) & \textbf{2.50} (0.28) & \textbf{3.05} (0.16) & \textbf{3.30} (0.03) & 2.89 (0.55) & \textbf{2.96} (0.51) & \textbf{3.49} (0.63) & \textbf{3.72} (0.27) \\
\bottomrule
    \end{tabular}
\end{table*}
}

\subsection{Occlusion or Open-space}
Since our framework relies on egocentric LiDAR scans, spatial regions situated behind static obstacles or pedestrians suffer from severe occlusions. Estimating motion dynamics behind opaque structures (e.g., solid walls) inherently relies on spatial extrapolation from visible layout geometry and is therefore unreliable in unseen environments. In contrast, dynamic pedestrian occlusions allow partial geometric information to pass through spatial gaps up to a certain crowd density, though progressive point cloud degradation can still impair prediction accuracy.

To isolate the impact of dynamic pedestrian occlusions, we analyze frame-by-frame NLL fluctuations corresponding to varying pedestrian configurations using stationary robot test sequences. Evaluated at identical physical locations, the model exhibits no systematic NLL degradation across varying pedestrian counts or occlusion levels. This structural robustness is largely attributable to our sequence-level training formulation: by aggregating observed motion samples across an entire sequence into a shared spatial target, the network effectively learns representations resilient to transient point-cloud shadows. Furthermore, comparative evaluations show that joint training with spatial frequency score maps substantially reduces worst-case NLLs, while Grid Sampling (GS) consistently improves estimation quality in heavily occluded frames.
\begin{figure}
    \centering
    \includegraphics[width=1\linewidth]{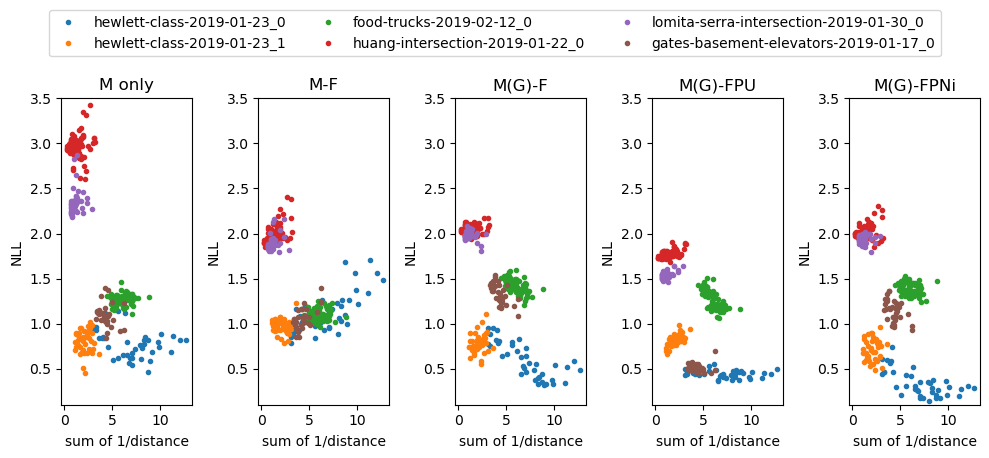}
    \caption{Occlusion sensitivity analysis. These plot show NLL for each frame in static-robot sequences from JRDB.}
    \label{fig:occlusion}
\end{figure}

\begin{figure}
    \centering
    \includegraphics[width=1\linewidth]{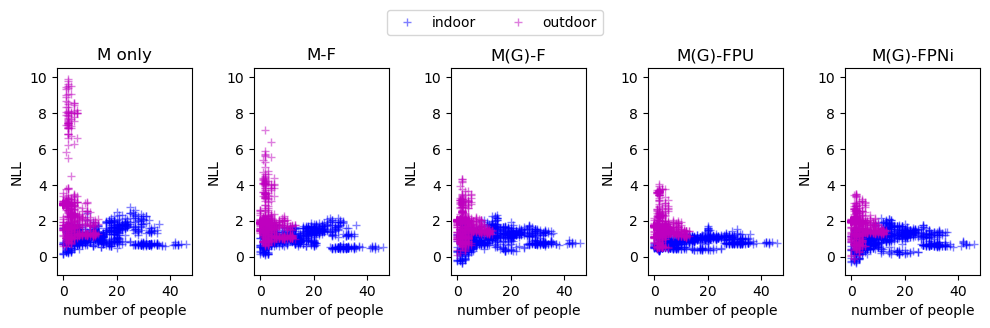}
    \caption{Indoor vs outdoor performance. These plots show NLL for each frame in test sequences from JRDB.}
    \label{fig:indoor_ouotdoor}
\end{figure}

Interestingly, worst NLLs occurred in low-density sequences, which predominantly feature spatially unconstrained outdoor environments as shown in \cref{fig:indoor_ouotdoor}, outdoor motion field estimation is more challenging than indoor estimation. This performance gap stems from two primary factors: first, underlying pedestrian motion distributions are inherently more diffuse in wide-open spaces; second, our model restricts point cloud inputs to a fixed spatial cropping range around the robot, thereby omitting critical distal geometric features such as distant building entrances or crosswalks. For downstream trajectory prediction, environment-dependent spatial priors exert significantly less constraint on pedestrian movement in expansive areas. Consequently, dynamically attenuating the influence of the MoD prior in highly unconstrained environments serves as an effective mechanism to mitigate performance degradation.

\section{Benchmark details}\label{sec:bench_detail}
\subsection{Occupancy map comparison}
Bayesian Floor Field (BFF)~\cite{francesco2024bff}, the primary existing transferable MoD baseline cannot be compared directly with our method in likelihood because BFF predicts discrete motion directions on a discrete spatial grid, whereas our framework outputs continuous velocity distributions parameterized by SWGMMs across continuous space. To establish a fair comparative baseline, we extend the BFF to output continuous SWGMM representations. Instead of 3D point clouds, this continuous variant \textit{BFF*} takes a 2D occupancy map as input, constructed by filtering the point clouds by height (removing ground and ceiling planes) and projecting the remaining points onto a bird's-eye-view grid.

As shown in \cref{tab:input}, while BFF* yields marginally superior performance in known environments, our proposed 3D LiDAR-based framework integrated with our training strategies consistently outperforms it in unknown environments. Furthermore, the results demonstrate that our proposed training techniques are equally effective when applied to the 2D input baseline, highlighting their general utility across different input representations.
\begin{table*}
\centering
\caption{Input type comparison. \textit{2D occ} refers to models that take occupancy images as input following BFF~\cite{francesco2024bff}. \textit{Motion} corresponds to the baseline that simply extends NeMo-map~\cite{zhu2026nemomap}. Each value represents the mean of three trials, with the standard deviation shown in parentheses. The lowest mean value per column is in bold.}
\label{tab:input}
\begin{tabular}{lccccccc}
\toprule
 & & \multicolumn{2}{c}{JRDB-Known} & \multicolumn{2}{c}{JRDB-Unknown} & \multicolumn{2}{c}{SiT} \\
Method & Input type & NLL & ($20^\circ$ var) & NLL & ($20^\circ$ var) & NLL & ($20^\circ$ var) \\
\midrule
BFF* & 2D occ & 1.17 (0.07) & 4.80 (0.62) & 1.53 (0.05) & 4.82 (0.14) & 1.34 (0.12) & 4.07 (1.64) \\
Motion & 3D pts & 1.22 (0.08) & 4.12 (0.62) & 1.79 (0.29) & 4.66 (0.35) & 1.63 (0.04) & 4.68 (0.86) \\
M(G)-FPNi & 2D occ & \textbf{1.05} (0.01) & 2.58 (0.25) & 1.43 (0.01) & 3.60 (0.17) & 1.35 (0.17) & 3.21 (0.52) \\
M(G)-FPNi & 3D pts & 1.12 (0.07) & \textbf{2.53} (0.27) & \textbf{1.42} (0.05) & \textbf{2.98} (0.51) & \textbf{1.24} (0.11) & \textbf{2.82} (0.74) \\
\bottomrule
    \end{tabular}
\end{table*}

\subsection{NeMo-map comparison}
NeMo-map~\cite{zhu2026nemomap} is not transferable and requires train data on the same location. To compare our transferable method and original NeMo-map on the same test data, we extracted overlapping pairs from JRDB train and test set. Note that not all sequences in "JRDB-Known" have clear spatial overlaps around sensor positions with train set. Three overlapping pairs are aligned with each other via occupancy map generated by LiDAR points and panoptic segmentation classes and then corrected manually. The temporal dependence i.e. SIREN in NeMo-map query encoding was removed here, to account for the temporally clustered set of training data per scene. 

\begin{figure*}
    \centering
    \includegraphics[width=0.33\linewidth, trim=40 0 40 0, clip]{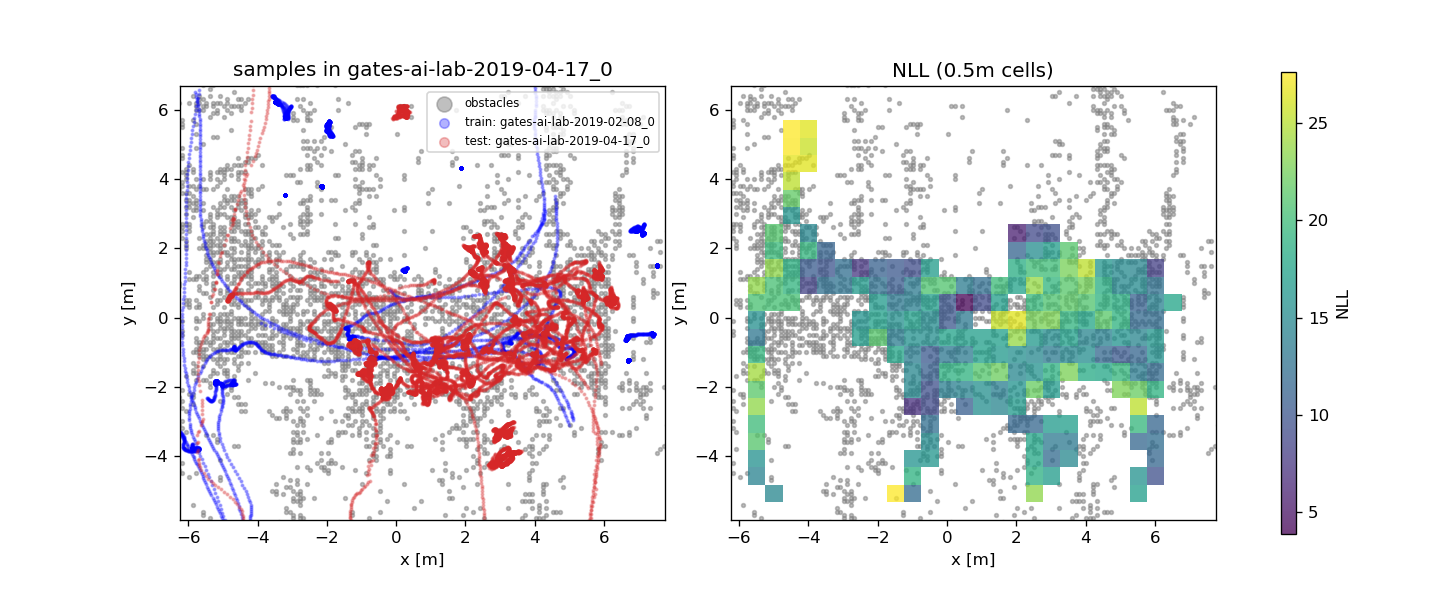}
    \includegraphics[width=0.33\linewidth, trim=40 0 40 0, clip]{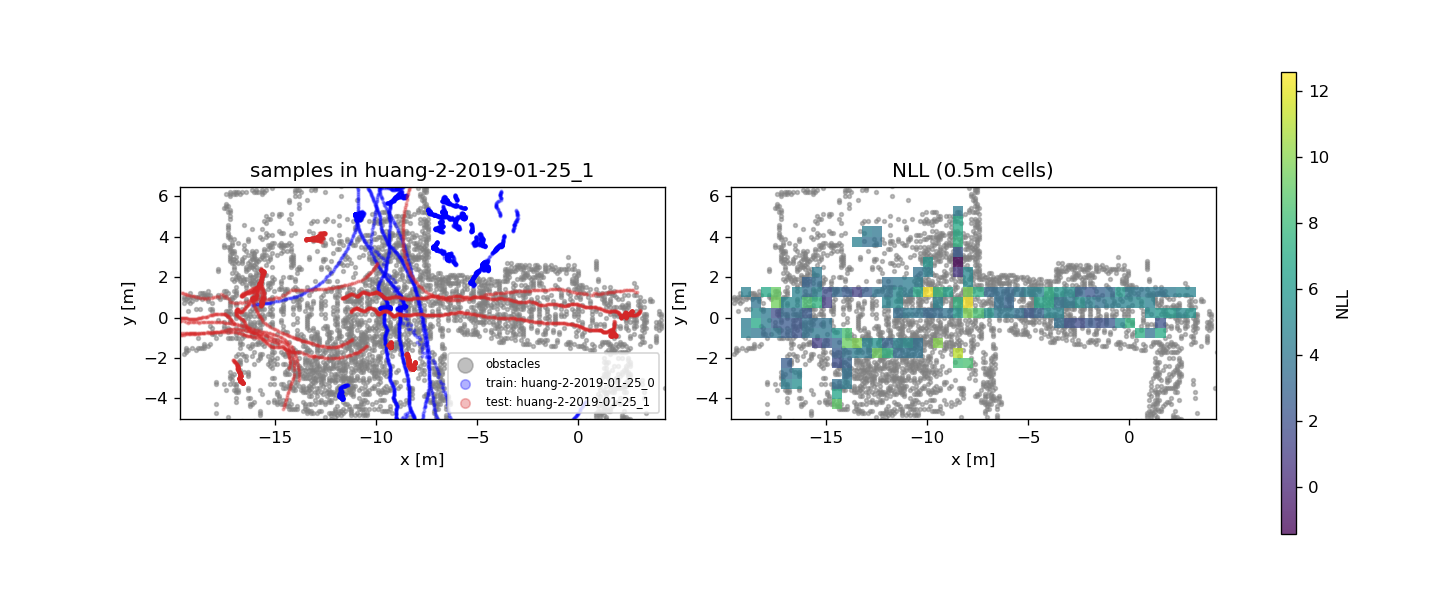}
    \includegraphics[width=0.33\linewidth, trim=40 0 40 0, clip]{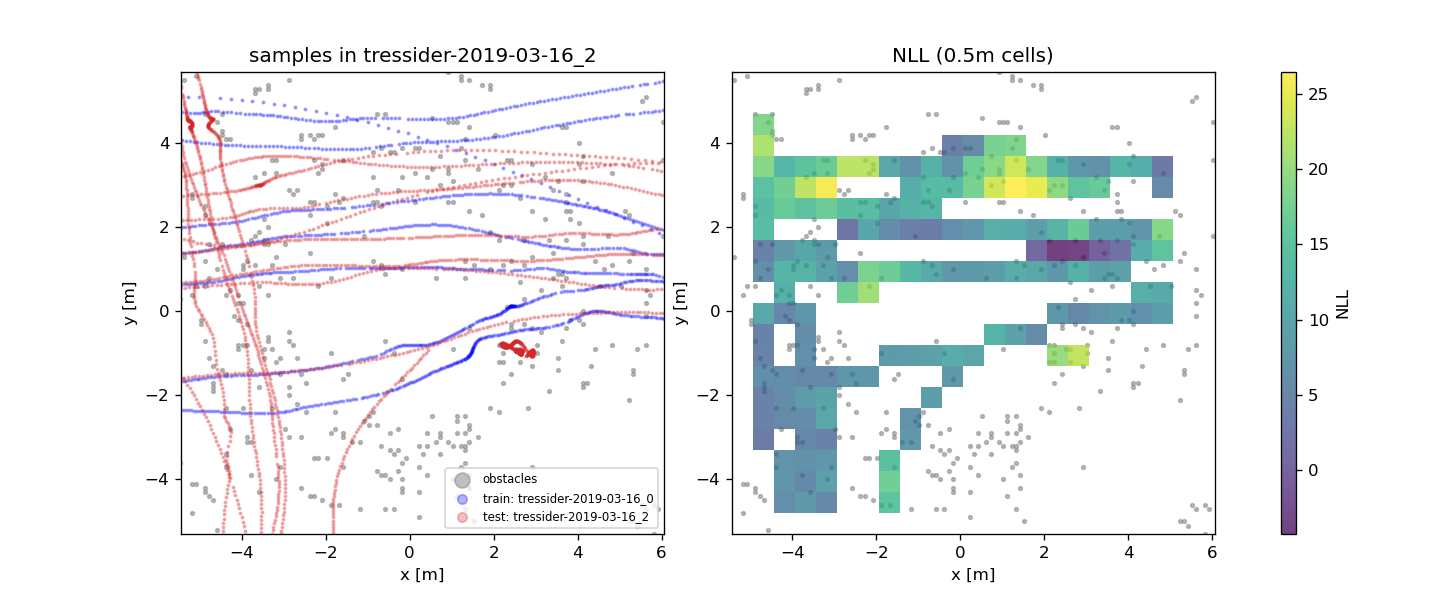}
    \caption{Known location evaluation for NeMo-map. Gray dots are LiDAR points that correspond to obstacle classes in panoptic segmentation labels.}
    \label{fig:nemo_known_plot}
\end{figure*}

\Cref{tab:known_mod} shows the NLL comparison. \Cref{fig:nemo_known_plot} shows the spatial coverages by train and test observations and NLL at each test grids. Thanks to continuous query of positions, all methods can infer at any position even where there are no train observations at same position. NeMo-map suffers from severe data sparsity per location, transferable models leveraging multi-location training data (BFF* and Ours) achieve substantially lower NLL values even in this known locations.

\begin{table*}
    \centering
        \caption{Comparison with NeMo-map~\cite{zhu2026nemomap} and extended BFF~\cite{francesco2024bff} on Known sequences in JRDB.  \textit{2D occ} refers to models that take occupancy images as input following BFF~\cite{francesco2024bff}.}
    \label{tab:known_mod}
    \begin{tabular}{lccccccc}
    \toprule
 && \multicolumn{2}{c}{gates-ai-lab-2019-04-17\_0} & \multicolumn{2}{c}{huang-2-2019-01-25\_1} & \multicolumn{2}{c}{tressider-2019-03-16\_2}  \\
Model(/Train method) & Input & NLL & ($20^\circ$ var) & NLL & ($20^\circ$ var) & NLL & ($20^\circ$ var) \\
\midrule
NeMo-map & None & 16.0 (0.91) & 16.9 (0.90) & 4.74 (0.92) & 7.62 (0.14) & 11.0 (1.65) & 12.3 (1.22) \\
BFF*/Motion & 2D occ & \textbf{1.03} (0.08) & 4.93 (1.08) & 0.98 (0.04) & 4.07 (1.23) & 0.99 (0.04) & 4.14 (1.62) \\
Ours-base/Motion & 3D pts & 1.18 (0.03) & 3.86 (0.10) & 0.83 (0.19) & 2.94 (1.33) & 1.06 (0.04) & 2.81 (0.75) \\
BFF*/M(G)-FPNi & 2D occ & 1.05 (0.02) & \textbf{2.81} (0.02) & 0.75 (0.14) & 1.64 (0.31) & \textbf{0.85} (0.10) & 2.49 (1.09) \\
Ours/M(G)-FPNi & 3D pts & 1.80 (0.10) & 3.36 (0.26) & \textbf{0.66} (0.21) & \textbf{1.57} (0.77) & 0.88 (0.19) & \textbf{2.27} (1.20) \\
\bottomrule
    \end{tabular}
\end{table*}

\subsection{Additional qualitative results}
\begin{figure*}[tbp]
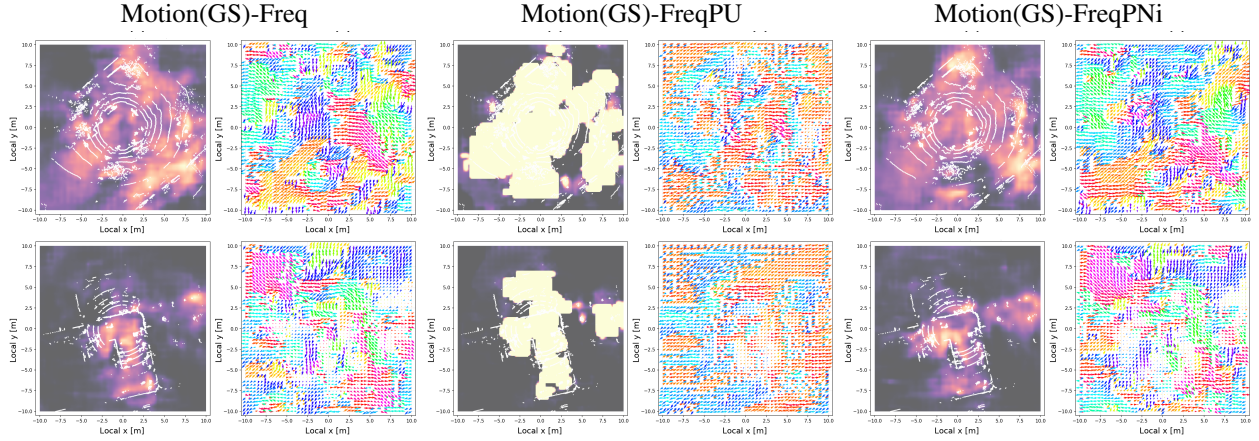

    \centering
    \begin{tabular}{c@{\hspace{3pt}}c@{\hspace{3pt}}c}
    Motion(GS)-Freq & Motion(GS)-FreqPU & Motion(GS)-FreqPNi \\
    \includegraphics[width=0.31\linewidth, trim=0 880 0 1750, clip]{figures/visualization/nemo_bce_mgs/test-latest-unseen_predictions_visualization_crop_edited.png} &
    \includegraphics[width=0.31\linewidth, trim=0 880 0 1750, clip]{figures/visualization/nemo_bcePU_mgs/test-latest-unseen_predictions_visualization_crop_edited.png} &
    \includegraphics[width=0.31\linewidth, trim=0 880 0 1750, clip]{figures/visualization/nemo_bcePNi_mgs/test-latest-unseen_predictions_visualization_crop_edited.png}
    \end{tabular}
    \vspace{-10pt}
    \caption{Frequency score and motion outputs of \textit{Motion(GS)-Freq}, \textit{Motion(GS)-FreqPU}, and \textit{Motion(GS)-FreqPNi}. For each method, the left panel shows the predicted frequency score, with brighter colors indicating higher scores, and the right panel shows the motion map. Arrow opacity, length, and hue represent mixture weight, speed, and heading direction, respectively.
    }
    \label{fig:freq_motion_outputs_2}
\end{figure*}

\Cref{fig:freq_motion_outputs_2} is the rest part of motion visualization in Fig.~5 in main paper.

\section{Trajectory Prediction}
\Cref{tab:longer_prediction} shows additional trajectory prediction results at long horizon up to $T=96$, which corresponds to $32$ seconds. The number of output modes is set to $3$. 
Unknown sequences show higher errors than Known sequences for both methods.
MoD-based frameworks are generally expected to show lower prediction errors over extended prediction horizons, where scene-dependent structural context increasingly dictates pedestrian motion. While ViTE achieves the lowest overall error across most metrics, our MoD method yields slightly lower errors in unseen environments at an extended prediction horizon of $T=96$. Nevertheless, because our proposed MoD estimator is spatially bounded to a local area around the instantaneous sensor location, extreme long-horizon predictions (note that this is out of scope in a social navigation application) are increasingly affected by boundary limitations, where queries beyond the observable spatial crop rely on zero-padded features.

\begin{table*}
    \centering
        \caption{Trajectory prediction errors on JRDB. All results are averaged over 3 training runs.}
    \label{tab:longer_prediction}
    \begin{tabular}{ll|cc|cc|cc|cc|cc}
    \toprule
  & & \multicolumn{2}{c}{T=12} & \multicolumn{2}{c}{T=18} & \multicolumn{2}{c}{T=24} & \multicolumn{2}{c}{T=48} & \multicolumn{2}{c}{T=96} \\
Split & Method & ADE & FDE & ADE & FDE & ADE & FDE & ADE & FDE & ADE & FDE \\
    \midrule
\multirow{2}{8em}{JRDB-Known} & ViTE & \textbf{0.28} & \textbf{0.46} & \textbf{0.41} & \textbf{0.67} & \textbf{0.53} & \textbf{0.88} & \textbf{0.91} & \textbf{1.59} & \textbf{1.29} & \textbf{2.32} \\
 & Ours & 0.31 & 0.58 & 0.45 & 0.84 & 0.57 & 1.09 & 0.98 & 1.83 & 1.63 & 2.70 \\
    \midrule
\multirow{2}{8em}{JRDB-Unknown} & ViTE & \textbf{0.42} & \textbf{0.65} & \textbf{0.61} & \textbf{0.98} & \textbf{0.79} & \textbf{1.30} & \textbf{1.47} & \textbf{2.64} & 2.46 & 4.56 \\
 & Ours & 0.47 & 0.89 & 0.69 & 1.35 & 0.89 & 1.77 & 1.53 & 3.10 & \textbf{2.36} & \textbf{4.52} \\
    \bottomrule
    \end{tabular}
\end{table*}

    
\end{document}